\documentclass[11pt]{article}

\newcommand{\confversion}[1]{}
\newcommand{\fullversion}[1]{#1}

\usepackage{hyperref}
\usepackage{url}
\usepackage{multirow}
\usepackage{stackrel}
\usepackage{comment}

\usepackage{amsfonts,epsfig,graphicx}
\usepackage{amsmath,amssymb,amsthm}
\usepackage{fullpage}
\usepackage{hhline}
\usepackage{sidecap}

\usepackage{epsf}
\usepackage{graphics}
\usepackage{amsfonts}
\usepackage{amsmath}
\usepackage{amsthm}
\usepackage{psfrag}

\usepackage{enumitem}
\setlist{itemsep=1pt, topsep=1pt}

\usepackage{caption}
\usepackage{subcaption}

\usepackage{xcolor,colortbl}

\usepackage{wrapfig}
\usepackage{fullpage}
\usepackage{picinpar}

\let\oldbibliography\thebibliography
\renewcommand{\thebibliography}[1]{%
  \oldbibliography{#1}%
  \setlength{\itemsep}{0pt}%
}
\newtheorem{theorem}{Theorem} 
 
\newtheorem{lemma}{Lemma}
\newtheorem{proposition}{Proposition}

\newcommand{\delcrit}{\ensuremath{\delta_n}}

\newcommand{\DelHat}{\ensuremath{\widehat{\Delta}}}

\newcommand{\tracer}[2]{\ensuremath{\langle \!\langle {#1}, \; {#2}
\rangle \!\rangle}}

\newcommand{\defn}{\ensuremath{:\,=}}
\newcommand{\Lnorm}[2]{\ensuremath{\|{#1}\|_{#2}}}

\newcommand{\Exs}{\ensuremath{\mathbb{E}}}
\newcommand{\argmax}{\operatornamewithlimits{arg~max}}
\newcommand{\argmin}{\operatornamewithlimits{arg~min}}

\newcommand{\kl}[2]{\ensuremath{D_{\mathrm{KL}}(#1\|#2)}}
\newcommand{\reals}{\ensuremath{\mathbb{R}}}
\newcommand{\mprob}{\ensuremath{\mathbb{P}}}

\newcommand{\half}{\ensuremath{{\frac{1}{2}}}}

\newcommand{\noise}{\ensuremath{\Wmat}}
\newcommand{\covnum}{\ensuremath{N}}
\newcommand{\metent}{\ensuremath{\log \covnum}}
\newcommand{\wt}{\ensuremath{M}}
\newcommand{\wtstar}{\ensuremath{\wt^*}}

\newcommand{\wthat}{\ensuremath{\widehat{\wt}}}
\newcommand{\wttil}{\ensuremath{\widetilde{\wt}}}

\newcommand{\numitems}{\ensuremath{n}}
\newcommand{\obs}{\ensuremath{Y}}

\newcommand{\perm}{\ensuremath{\pi}}
\newcommand{\permhat}{\ensuremath{\hat{\perm}}}

\newcommand{\permhatclass}{\ensuremath{\widehat{\Pi}}}

\newcommand{\permid}{{\mbox{\scriptsize id}}}

\newcommand{\chatterjeeclass}{\ensuremath{\mathbb{C}_{\mbox{\scalebox{.5}{{SST}}}}}}

\newcommand{\chattstar}{\ensuremath{\widetilde{\mathbb{C}}_{\mbox{\scalebox{.5}{{SST}}}}}}

\newcommand{\chattperm}[2][]{\ensuremath{#1{\mathbb{C}}_{\mbox{\scalebox{.5}{{SST}}}}(#2)}}
\newcommand{\bisoclassminus}[1]{\ensuremath{\chatterjeeclass(#1; [-1,1])}}

\newcommand{\plaincon}{c}

\newcommand{\packnum}{\ensuremath{T}}

 \newcommand{\ones}{\ensuremath{1}}

\newcommand{\factorial}[1]{\ensuremath{#1!}}

\newcommand{\AuxEvent}{\ensuremath{\mathcal{A}_t}}

\newcommand{\numobs}{\numitems}


\newcounter{parentnumber}

\newcommand{\chattDiffmx}{\ensuremath{D}}
\newcommand{\chattDiff}{\ensuremath{\mathbb{C}_{\mbox{\scalebox{.5}{{DIFF}}}}}}

\newcommand{\matsnorm}[2]{|\!|\!| #1 | \! | \!|_{{#2}}}

\newcommand{\numitem}{\ensuremath{n}}
\newcommand{\order}{\ensuremath{\mathcal{O}}}

\long\def\comment#1{}

\newcommand{\frobnorm}[1]{\ensuremath{\matsnorm{#1}{\mbox{\tiny{F}}}}}

\newcommand{\Mhat}{\ensuremath{\widehat{M}}}
\newcommand{\Mstar}{\ensuremath{M^*}}

\newcommand{\Wmat}{\ensuremath{W}}

\newcommand{\real}{\ensuremath{\mathbb{R}}}

 \newcommand{\pistar}{\pi^*}
\newcommand{\trace}{\operatorname{trace}}

\newenvironment{carlist} {\begin{list}{$\bullet$}
    {\setlength{\topsep}{0in} \setlength{\partopsep}{0in}
      \setlength{\parsep}{0in} \setlength{\itemsep}{\parskip}
      \setlength{\leftmargin}{0.07in} \setlength{\rightmargin}{0.08in}
      \setlength{\listparindent}{0in} \setlength{\labelwidth}{0.08in}
      \setlength{\labelsep}{0.1in} \setlength{\itemindent}{0in}}}
               {\end{list}}

\newcommand{\bcar}{\begin{carlist}} \newcommand{\ecar}{\end{carlist}}

\newcommand{\KCON}{\ensuremath{c}}
\newcommand{\ULOW}{\ensuremath{\KCON_\ell}}
\newcommand{\UUP}{\ensuremath{\KCON_u}}
\newcommand{\UHP}{\ensuremath{\KCON}}

\newcommand{\Mtil}{\ensuremath{\widetilde{M}}}

\newcommand{\Win}{\ensuremath{N}} \newcommand{\Mbar}{\ensuremath{M_0}}

\newcommand{\projunit}{\ensuremath{\Pi_{[0,1]}}}

\newcommand{\numcomm}{s} \newcommand{\commsize}{k}
\newcommand{\commmax}{k_{\max}}
\newcommand{\commset}{\mathbf{\commsize}}
\newcommand{\wtlse}{\wthat_{LS}}
\newcommand{\chatterjeeclassPC}{\widetilde{\mathbb{C}}_{\mbox{\tiny
      SST}}} 

\newcommand{\AEsym}{\alpha_\numobs}
\newcommand{\wtCRL}{\ensuremath{{\Mhat}_{\mbox{\tiny{CRL}}}}}

\newcommand{\permcrl}{\perm_{\mbox{\tiny CRL}}}

\newcommand{\planted}{\kappa}
\newcommand{\graph}{\mathcal{G}}

\newcommand{\MYSET}{\ensuremath{P}}
\newcommand{\ORARISK}{R_\numitems}

\newcommand{\reg}[1]{\ensuremath{(\numitems - #1 + 1) (\log \numitems)^3}}
\newcommand{\regsimple}[1]{\ensuremath{- #1(\log \numitems)^3}}
\newcommand{\wtreg}{\ensuremath{\widehat{\wt}_{\mbox{\tiny REG}}}}
\newcommand{\wtcomm}{\ensuremath{\wthat_\commsize}}
\newcommand{\maxsizefn}{\ensuremath{\commmax}}

\newcommand{\adaptZ}{\ensuremath{\zeta}}

\newcommand{\commstar}{\ensuremath{\commsize^*}}

\newcommand{\classgenup}{\mathbb{C}}
\newcommand{\wthatgenup}{\wthat}
\newcommand{\reggenup}{\lambda}
\newcommand{\bgenup}{b}
\newcommand{\hgenup}{h}
\newcommand{\ggenup}{g}
\newcommand{\ballgenup}{t}
\newcommand{\tailgenup}{u}

\newcommand{\card}{\ensuremath{\operatorname{card}}}
\newcommand{\CODE}{\ensuremath{\mathbb{B}}}
\newcommand{\MZERO}{\ensuremath{M(\mathbf{0})}}
\newcommand{\MZEROSUB}[1]{\ensuremath{M_{#1}(\mathbf{0})}}
\newcommand{\cword}{\ensuremath{\mathbf{z}}} 
\newcommand{\HACKPERM}{P_\pi}
\newcommand{\LINE}{\ensuremath{\mathbb{L}(\wtstar, \wtlse)}}
\newcommand{\commax}{\commmax}

\begin{document}

\title{Adaptivity in Estimation from Paired Comparisons} 

\title{{\bf{Feeling the Bern: Adaptive Estimators for \\Bernoulli
  Probabilities of Pairwise Comparisons}}}

\author{
{\large{
\begin{tabular}{ccc}
Nihar  B. Shah & Sivaraman Balakrishnan & Martin J. Wainwright\\
\small Dept. of EECS & \small Dept. of Statistics & \small Dept. of EECS and Statistics\\
\small UC Berkeley & \small CMU & \small UC Berkeley \\
 \small nihar@eecs.berkeley.edu &  \small siva@stat.cmu.edu &  \small wainwrig@berkeley.edu
\end{tabular}
}}
}

\date{}

\maketitle
\thispagestyle{empty}

\begin{abstract}
We study methods for aggregating pairwise comparison data in order to estimate outcome probabilities for future comparisons among a collection of $\numitems$ items. Working within a flexible framework that imposes only a form of strong stochastic transitivity (SST), we introduce an adaptivity index defined by the indifference sets of the pairwise comparison probabilities. In addition to measuring the usual worst-case risk of an estimator, this adaptivity index also captures the extent to which the estimator adapts to instance-specific difficulty relative to an oracle estimator. We prove three main results that involve this adaptivity index and different algorithms. First, we propose a three-step estimator termed Count-Randomize-Least squares (CRL), and show that it has adaptivity index upper bounded as $\sqrt{\numitems}$ up to logarithmic factors. We then show that that conditional on the hardness of planted clique, no computationally efficient estimator can achieve an adaptivity index smaller than $\sqrt{\numitems}$. Second, we show that a regularized least squares estimator can achieve a poly-logarithmic adaptivity index, thereby demonstrating a $\sqrt{\numitems}$-gap between optimal and computationally achievable adaptivity. Finally, we prove that the standard least squares estimator, which is known to be optimally adaptive in several closely related problems, fails to adapt in the context of estimating pairwise probabilities.
\end{abstract}


\section{Introduction}

There is an extensive literature on modeling and analyzing data in the
form of pairwise comparisons between items, with much of the earliest
literature focusing on applications in voting, social choice theory,
and tournaments.  The advent of new internet-scale applications,
particularly search engine ranking~\cite{radlinski2008does}, online
gaming~\cite{herbrich2007trueskill}, and
crowdsourcing~\cite{shah2015estimation}, has renewed interest in
ranking problems, particularly in the statistical and computational
challenges that arise from the aggregation of large data sets of
paired comparisons.

The problem of aggregating pairwise comparisons, which may be
inconsistent and/or noisy, presents a number of core challenges,
including: (i) how to produce a consensus ranking from the paired
comparisons~\cite{braverman2008noisy, rajkumar2015ranking,
  shah2015simple}; (ii) how to estimate a notional ``quality'' for
each of the underlying objects~\cite{negahban2012iterative,
  hajek2014minimax, shah2015estimation}; and (iii) how to estimate the
probability of the outcomes of subsequent
comparisons~\cite{chatterjee2014matrix, shah2015stochastically}.  In
this paper, we focus on the third task---that is, the problem of
estimating the probability that one object is preferred to another.  Accurate knowledge of such pairwise comparison
probabilities is useful in various applications, including (in
operations research) estimating the probability of a customer picking
one product over another, or (in sports bookmaking and tournament
design) estimating the probability of one team beating another.

In more detail, given a set of $\numitems$ items
$\{1,\ldots,\numitems\}$, the paired comparison probabilities can be
described by an $(\numitems \times \numitems)$ matrix $\wtstar$ in
which the $(i,j)^{\mathrm{th}}$ entry corresponds to the probability
that item $i$ beats item $j$. From this perspective, problem of
estimating the comparison probabilities amounts to estimating the
unknown matrix $\wtstar$.  In practice, one expects that the pairwise
comparison probabilities exhibit some form of structure, and in this
paper, in line with some past work on the problem, we assume that the
entries of the matrix $\wtstar$ satisfy the strong stochastic
transitivity (SST) constraint.  It is important to note that the SST
constraint is considerably weaker than standard parametric assumptions
that are often made in the literature---for instance, that the entries
of $\wtstar$ follow a Bradley-Terry-Luce~\cite{bradley1952rank,
  luce1959individual} or Thurstone~\cite{thurstone1927law} model.  The
SST constraint is quite flexible and models satisfying this constraint
often provide excellent fits to paired comparison data in a variety of
applications. There is also a substantial body of empirical work that
validates the SST assumption---for instance, see the
papers~\cite{mclaughlin1965stochastic, davidson1959experimental,
  ballinger1997decisions} in the psychology and economics literatures.


On the theoretical front, some past work~\cite{chatterjee2014matrix,
  shah2015stochastically} has studied the problem of estimating SST
matrices in the Frobenius norm.  These works focus exclusively on the
\emph{global} minimax error, meaning that the performance of any
estimator is assessed in a worst-case sense globally over the entire
SST class. It is well-known that the criterion of global minimax can
lead to a poor understanding of an estimator, especially in situations
where the intrinsic difficulty of the estimation task is highly
variable over the parameter space (see, for instance, the discussion
and references in Donoho et al.~\cite{donoho1995wavelet}).  In such
situations, it can be fruitful to benchmark the risk of an estimator
against that of a so-called oracle estimator that is provided with
side-information about the local structure of the parameter space.
Such a benchmark can be used to show that a given estimator is
\emph{adaptive}, in the sense that even though it is not given
side-information about the problem instance, it is able to achieve
lower risk for ``easier'' problems (e.g., see the
papers~\cite{candes2006modern, koltchinskii2011oracle,
  cai2011framework} for results of this type).\footnote{The term
  ``adaptivity'' in this paper derives its meaning from the literature
  on statistics, and refers to the property of an estimator of
  automatically adapting its performance to the complexity of the
  problem. It should not be confused with the notion of ``adaptive
  sampling'' used in the context of sequential design or adaptive
  learning, which refers to the ability to obtain samples one at a
  time in a sequential and data-dependent manner.}  In this paper, we
study the problem-specific difficulty of estimating a pairwise
comparison matrix $\wtstar$ by introducing an adaptivity index that
involves the size of the indifference sets in the matrix $\wtstar$.
These indifference sets, which arise in many relevant applications,
correspond to subsets of items that are all equally desirable.

In addition, our work makes contributions to a growing body of work
(e.g.,~\cite{rigollet,ma2015computational, Wai14_ICM}) that studies the notion of
a computationally-constrained statistical risk.

\noindent In more detail, we make the following contributions in this
paper:
\begin{itemize}[leftmargin = *]
\item We show that the risk of estimating a pairwise comparison
  probability matrix $\wtstar$ depends strongly on the size of its
  largest indifference set.  This fact motivates us to define an
  adaptivity index that benchmarks the performance of an estimator
  relative to that of an oracle estimator that is given additional
  side information about the size of the indifference sets in
  $\wtstar$.  By definition, an estimator with lower values of this index is
  said to exhibit better adaptivity, and the oracle estimator has an adaptivity
  index of $1$.
\item We characterize the fundamental limits of adaptivity, in
  particular by proposing a regularized least squares estimator with a
  carefully chosen regularization function.  With a suitable choice of
  regularization parameter, we prove that this estimator achieves an
  $\widetilde{\order}(1)$ adaptivity index, which matches the best
  possible up to poly-logarithmic factors.
\item We then show that conditional on the planted clique hardness
  conjecture, the adaptivity index achieved by any polynomial-time
  algorithm must be lower bounded as
  $\widetilde{\Omega}(\sqrt{\numitems})$. This result exhibits an
  interesting gap between the adaptivity of polynomial-time versus
  statistically optimal estimators.
\item We propose a computationally-efficient three-step
  ``Count--Randomize--Least squares'' (CRL) estimator for estimation
  of SST matrices, and show that its adaptivity index is upper bounded
  as $\widetilde{\order}(\sqrt{\numitems})$. Due to the aforementioned
  lower bound, the CRL estimator has the best possible adaptivity
  among all possible computationally efficient estimators.
\item Finally, we investigate the adaptivity of the standard
  (unregularized) least squares estimator. This estimator is found to
  have good, or even optimal adaptivity in several related problems,
  and is also minimax-optimal for the problem of estimating SST
  matrices. We prove that surprisingly, the adaptivity of the least
  squares estimator for estimating SST matrices is of the order
  $\widetilde{\Theta}(\numitems)$, which is as bad as a constant estimator
  that is independent of the data.
\end{itemize}

The remainder of this paper is organized as follows.  We begin in
Section~\ref{SecSetting} with background on the problem.
Section~\ref{SecMainResults} is devoted to the statement of our main
results, as well as discussion of their consequences.  In
Section~\ref{SecProofs}, we provide the proofs of our main results,
with the more technical details deferred to appendices. Finally, Section~\ref{SecDiscussion} presents concluding remarks.


\section{Background and problem setting}
\label{SecSetting}

In this section, we provide background and a more precise problem
statement.


\subsection{Estimation from pairwise comparisons}
\label{SecSettingPairwise}

Given a collection of $\numitem$ items, suppose that we arrange the
paired comparison probabilities in a matrix \mbox{$\wtstar \in
  [0,1]^{\numitem \times \numitem}$,} where $\Mstar_{ij}$ is the
probability that item $i$ is preferred to item $j$ in a paired
comparison.  Accordingly, the upper and lower halves of $\Mstar$ are
related by the \fullversion{shifted-skew-symmetry} condition
$\Mstar_{ji} = 1 - \Mstar_{ij}$ for all $i, j \in [\numitems]$, where
we assume that $\Mstar_{ii} = 0.5$ for all $i \in [\numitems]$ for
concreteness. In other words, the shifted matrix $\Mstar - \frac{1}{2}
\ones \ones^T$ is skew-symmetric.  Here we have adopted the standard
shorthand $[\numitems] \defn \{1, 2, \ldots, \numitems \}$.

Suppose that we observe a random matrix \mbox{$Y \in \{0,
  1\}^{\numitems \times \numitems}$} with (upper-triangular)
independent Bernoulli entries, in particular, with 
\begin{align}
\label{EqnDefnBernModel}
\mathbb{P}[Y_{ij}
  = 1] = \wtstar_{ij} \qquad \mbox{ for every $1 \leq i \leq j \leq n$},
  \end{align}
 and $Y_{ji}
= 1 - Y_{ij}$ except on the diagonal. We take the diagonal entries
$Y_{ii}$ to be $\{0,1\}$ with equal probability, for every $i \in
[n]$.  The focus of this work is not to evaluate the effects of the
choice of the pairs compared, but to understand the effects of the
noise models. Consequently, we restrict attention to the case of a
single observation per pair, but keeping mind in that one may extend
the result to other observation models via techniques similar to those
proposed in our past work~\cite{shah2015estimation,
  shah2015stochastically}.  Based on observing $Y$, our goal in this
paper is to recover an accurate estimate, in the squared Frobenius
norm, of the full matrix $\wtstar$.

We consider matrices $\wtstar$ that satisfy the constraint of
\emph{strong stochastic transitivity} (SST), which reflects the
natural transitivity of any complete ordering.  Formally, suppose that
the set of all items $[\numitem]$ is endowed with a complete ordering
$\pistar$.  We use the notation \mbox{$\pistar(i) \succ \pistar(j)$}
to indicate that item $i$ is preferred to item $j$ in the total
ordering $\pistar$.  We say that the $\Mstar$ satisfies the SST
condition with respect to the permutation $\pistar$---or is
$\pistar$-SST for short---if
\begin{align}
\label{EqnTripleSST}
\Mstar_{ik} \geq \Mstar_{jk} \qquad \mbox{for every triple $(i,j,k)$
  such that $\pistar(i) \succ \pistar(j)$.}
\end{align}
The intuition underlying this constraint is as follows: since $i$
dominates $j$ in the true underlying order, when we make noisy
comparisons, the probability that $i$ is preferred to $k$ should be at
least as large as the probability that $j$ is preferred to $k$.  The
class of all SST matrices is given by
\begin{align}
\label{EqnDefnSST}
\chatterjeeclass & \defn \Big \{ M \in [0,1]^{\numitem \times
  \numitem} \, \mid \, \mbox{ $M_{ba} = 1 - M_{ab} \; \forall \;
  (a,b)$ and $\exists ~ \pi$ such that $M$ is $\pi$-SST} \Big \}.
\end{align}
The goal of this paper\footnote{We note that an accurate estimate of
  $\wtstar$ leads to an accurate estimate of the underlying
  permutation as well~\cite{shah2015stochastically}.}
is to design estimators that can estimate the true underlying matrix
\mbox{$\wtstar \in \chatterjeeclass$} from the observed matrix $\obs$.


\subsection{Indifference sets}

We now turn to the notion of \emph{indifference sets}, which allows
for a finer-grained characterization of the difficulty of estimating a
particular matrix. Suppose that the set $[\numitems]$ of all items is
partitioned into the union of $\numcomm$ disjoint sets
$\{\MYSET_i\}_{i=1}^\numcomm$ of sizes $\commset = (\commsize_1
,\ldots,\commsize_\numcomm)$ such that $\sum_{i=1}^{\numcomm}
\commsize_i = \numitems$. For reasons to be clarified in a moment, we
term each of these sets as an \emph{indifference set}.  We write $i \sim
i'$ to mean that the pair $i$ and $i'$ belong to the same index set,
and we say that a matrix $\wtstar \in \real^{\numitems \times
  \numitems}$ \emph{respects the indifference set partition}
$\{\MYSET_i\}_{i=1}^\numcomm$ if
\begin{align}
\wtstar_{ij} = \wtstar_{i' j'} \quad \mbox{for all quadruples $(i,j,
  i', j')$ such that $i \sim i'$ and $j \sim j'$}.
\end{align}
For instance, in the special case of a two-contiguous-block partition,
the matrix $\wtstar$ must have a $(2 \times 2)$ block structure, with
all entries equaling $1/2$ in the two diagonal blocks, all entries equaling $\alpha \in [0,1]$
in the upper right block, and equaling $(1-\alpha)$ in the lower left block.
Intuitively, matrices with this type of block structure should be
easier to estimate.

Indifference sets arise in various applications of ranking: for
instance, in buying cars, frugal customers may be indifferent between
high-priced cars; or in ranking news items, people from a certain
country may be indifferent to the domestic news from other countries.
Block structures of this type are also studied in other types of
matrix estimation problems, in which contexts they have been termed
communities, blocks, or level sets, depending on the application under
consideration. For instance, see the papers~\cite{abbe2015recovering,
  ma2015computational, chatterjee2015biisotonic} as well as references
therein for more discussion in such structures.

Given the number of partitions $\numcomm$ and their size vector
$\commset = (\commsize_1, \ldots, \commsize_\numcomm)$, we let
$\chatterjeeclass(\numcomm,\commset)$ denote the subset of
$\chatterjeeclass$ comprising all SST matrices that respect some
indifference set partition $\{\MYSET_i\}_{i=1}^\numcomm$ of sizes
$\commset$. The size of the largest indifference set $\commmax \defn
\Lnorm{\commset}{\infty} = \max \limits_{ i \in \{1,\ldots, s\}}
\commsize_i$ plays an important role in our analysis.  We
also use the notation $\chatterjeeclass(\commmax)$ to denote all SST
matrices that have at least one indifference set of size at least
$\commmax$, that is,
\begin{align*}
\chatterjeeclass(\commmax) \defn \bigcup_{ \|\commset\|_\infty \geq
  \commmax} \chatterjeeclass(\numcomm,\commset ),.
\end{align*}
Finally, with a minor abuse of notation, for any matrix
$\wt \in \chatterjeeclass$, we let $\maxsizefn(\wt)$ denote the size
of the largest indifference set in $\wt$.


\subsection{An oracle estimator and the adaptivity index}
\label{SecSettingAE}

We begin by defining a benchmark based on the performance of the best
estimator that has side-information that $\wtstar \in
\chatterjeeclass(\numcomm, \commset)$, along with the values of
$(\numcomm, \commset)$.  We evaluate any such estimator
$\Mtil(\numcomm, \commset)$ based on its mean-squared Frobenius
error
\begin{align}
\Exs[ \frobnorm{ \wttil(\numcomm,\commset) - \wtstar }^2 ] & = \Exs \Big[
  \sum_{i,j=1}^\numitems \big ( \wttil_{ij}(\numcomm, \commset) -
  \wtstar_{ij} \big)^2 \Big],
\end{align}
where the expectation is taken with respect to the random matrix $\obs
\in \{0,1\}^{\numitems \times \numitems}$ of noisy comparisons.  With
this notation, the \emph{$(\numcomm, \commset)$-oracle risk} is given
by
\begin{align}
\label{EqnOracleRisk}
\ORARISK(\numcomm, \commset) & \defn \inf \limits_{\wttil(
  \numcomm,\commset)} \sup \limits_{\wtstar \in
  \chatterjeeclass(\numcomm,\commset)} \Exs[ \frobnorm{
    \wttil(\numcomm,\commset) - \wtstar }^2 ],
\end{align}
where the infimum is taken over all measurable functions
$\wttil(\numcomm, \commset)$ of the data $Y$.

For a given estimator $\wthat$ that \emph{does not know} the values of
$(\numcomm, \commset)$, we can then compare its performance to this
benchmark via the \emph{$(\numcomm, \commset)$-adaptivity index}
\begin{subequations}
\begin{align}
\label{EqnAdaptivity}
\AEsym(\wthat; \numcomm, \commset) \defn \frac{ \sup \limits_{\wtstar
    \in \chatterjeeclass(\numcomm,\commset)} \Exs[ \frobnorm{\wthat -
      \wtstar}^2] }{ \ORARISK(\numcomm, \commset)}.
\end{align}
The \emph{global adaptivity index} $\AEsym(\wthat)$ of an estimator
$\wthat$ is then given by
\begin{align}
\label{EqnGlobalAdaptivity}
\AEsym(\wthat) & \defn \max_{\numcomm, \commset:
  \Lnorm{\commset}{\infty} < \numitems} \AEsym(\wthat; \numcomm,
\commset).
\end{align}
\end{subequations}
In this definition, we restrict the maximum to the interval
$\Lnorm{\commset}{\infty} < \numitems$ since in the (degenerate) case
of $\Lnorm{\commset}{\infty} = \numitems$, the only valid matrix
$\wtstar$ is the all-half matrix and hence the estimator with the
knowledge of the parameters trivially incurs an error of zero.

Given these definitions, the goal is to construct estimators that are
computable in polynomial time, and possess a low adaptivity index.
Finally, we note that an estimator with a low adaptivity index also
achieves a good worst-case risk: any estimator $\wthat$ with global
adaptivity index $\AEsym(\wthat) \leq \gamma$ is minimax-optimal
within a factor $\gamma$.



\section{Main results}
\label{SecMainResults}

In this section, we present the main results of this paper on both
statistical and computational aspects of the adaptivity index.  We
begin with an auxiliary result on the risk of the oracle estimator
which is useful for our subsequent analysis.


\subsection{Risk of the oracle estimator}

Recall from Section~\ref{SecSettingAE} that the oracle estimator has
access to additional side information on the values of the number
$\numcomm$ and the sizes $\commset =
(\commsize_1,\ldots,\commsize_\numcomm)$ of the indifference sets of
the true underlying matrix $\wtstar$. The oracle estimator is defined
as the estimator that incurs the lowest possible
risk~\eqref{EqnOracleRisk} among all such estimators.

\noindent The following result provides tight bounds on the risk of
the oracle estimator.
\begin{proposition}
\label{PropORA}
There are positive universal constants $\ULOW$ and $\UUP$ such that
the $(\numcomm, \commset)$-oracle risk~\eqref{EqnOracleRisk} is
sandwiched as
\begin{align}
\label{EqnORAsandwich}
\ULOW (\numitems - \commax) \leq \ORARISK(\numcomm, \commset) \leq
\UUP(\numitems - \commmax + 1)(\log \numitems)^2.
\end{align}
\end{proposition}
\noindent Proposition~\ref{PropORA} provides a characterization
minimax risk of estimation under various subclasses of
$\chatterjeeclass$.
Remarkably, the minimax risk depends on only the size \mbox{$\commmax
  \defn \Lnorm{\commset}{\infty}$} of the largest indifference set:
given this value, it is not affected by the number of indifference
sets $\numcomm$ nor their sizes $\commset$. This property is in sharp
contrast to known results~\cite{chatterjee2015biisotonic} for the related problem of bivariate isotonic regression, in which the number $\numcomm$ of
indifference sets does play a strong role.

Note that when $\commax < \numitems$, we have $\half(\numitems - \commax + 1) \leq (\numitems - \commax)$, and consequently the lower bound in~\eqref{EqnORAsandwich} can be replaced by $\frac{\ULOW}{2}(\numitems - \commax + 1)$.

\subsection{Fundamental limits on adaptivity}
\label{SecStat}

Proposition~\ref{PropORA} provides a sharp characterization of the
denominator in the adaptivity index~\eqref{EqnAdaptivity}.  In this
section, we investigate the fundamental limits of adaptivity by
studying the numerator but disregarding computational constraints.
The main result of this section is to show that a suitably regularized
form of least-squares estimation has optimal adaptivity up to
logarithmic factors.

More precisely, recall that $\maxsizefn(\wt)$ denotes the size of the
largest indifference set in the matrix $\wt$. Given the observed
matrix $\obs$, consider the $M$-estimator
\begin{align}
\label{EqnDefnRegLse}
\wtreg \in \argmin \limits_{\wt \in \chatterjeeclass} \Big( \frobnorm{
  \wt - \obs }^2 \regsimple{\maxsizefn(\wt)} \Big).
\end{align}
Here the inclusion of term $-\maxsizefn(\wt)$, along with its
logarithmic weight, serves to ``reward'' the estimator for returning a
matrix with a relatively large maximum indifference set.  As our later
analysis in Section~\ref{SecLSEbad} will clarify, the inclusion of
this term is essential: the unregularized form of least-squares has
very poor adaptivity properties.

The following theorem provides an upper bound on the estimation error
and the adaptivity of the estimator $\wtreg$.
\begin{theorem}
\label{ThmRegLse}
There are universal constants $\UUP$ and $\UUP'$ such that for every
$\wtstar \in \chatterjeeclass$, the regularized least squares
estimator~\eqref{EqnDefnRegLse} has squared Frobenius error at most
\begin{subequations}
\begin{align}
\label{EqnRegLseUpper}
\frac{1}{\numitems^2} \frobnorm{ \wtstar - \wtreg}^2 \leq \UUP
\frac{\numitems - \maxsizefn(\wtstar) + 1}{\numitems^2} \; (\log
\numitems)^3,
\end{align}
with probability at least $1 - e^{- \half (\log \numitems)^2}$.
Consequently, its adaptivity index is upper bounded as
\begin{align}
\label{EqnRegLseAdaptive}
\AEsym(\wtreg) \leq \UUP' (\log \numitems)^3.
\end{align}
\end{subequations}
\end{theorem}
Since the adaptivity index of any estimator is at least $1$ by
definition, we conclude that the regularized least squares estimator
$\wtreg$ is optimal up to logarithmic factors.

The reader may notice that the optimization
problem~\eqref{EqnDefnRegLse} defining the regularized least squares
estimator $\wtreg$ is non-trivial to solve; it involves both a
nonconvex regularizer, as well as a nonconvex constraint set.  We shed
light on the intrinsic complexity of computing this estimator in 
Section~\ref{SecLower}, where we investigate the adaptivity index
achievable by estimators that are computable in polynomial time.


\subsection{Adaptivity of the CRL estimator}
\label{SecComp}

In this section, we propose a polynomial-time computable estimator termed the \emph{Count-Randomize-Least-Squares (CRL)} estimator, and prove an
upper bound on its adaptivity index.  In order to define the CRL
estimator, we requre some additional notation.  For any permutation
$\perm$ on $\numitems$ items, let $\chattperm{\perm} \subseteq
\chatterjeeclass$ denote the set of all SST matrices that are faithful
to the permutation $\perm$---that is
\begin{align}
\label{EqnDefnbisoclass}
\chattperm{\perm} \defn \big \{ \wt \in [0,1]^{\numitem \times
  \numitem} \, \mid \, \mbox{ $M_{ba} = 1 - M_{ab} \; \forall \;
  (a,b)$}, \; \mbox{$M_{ik} \geq M_{jk} \; \forall \; i,j,k \in
  [\numitems]$ s.t. $\perm(i) > \perm(j)$} \big \}.
\end{align}
One can verify that the sets $\{ \chattperm{\perm} \}$ for all
permutations $\perm$ on $\numitems$ items form a partition of the SST
class $\chatterjeeclass$.

The CRL estimator
acts on the observed matrix $\obs$ and outputs an estimate $\wtCRL \in
\chatterjeeclass$ via a three step procedure:\\
\underline{Step 1 (Count):} For each $i \in [\numitems]$, compute the
total number $\Win_i = \sum_{j = 1}^{\numitems} Y_{ij}$ of pairwise comparisons
that it wins.  Order the $\numitems$ items in terms of
$\{\Win_i\}_{i=1}^\numitems$, with ties broken arbitrarily. \\
\underline{Step 2 (Randomize):} Find the largest subset of items $S$
such that $|N_i - N_j| \leq \sqrt{\numitems} \log \numitems$ for
all $i, j \in S$.  Using the order computed in Step 1, permute this
(contiguous) subset of items uniformly at random within the
subset. Denote the resulting permutation as $\permcrl$.\\
\underline{Step 3 (Least squares):} Compute the least squares estimate
assuming that the permutation $\permcrl$ is the true permutation of
the items:
\begin{align}
\label{EqnDefnCRL}
\wtCRL \in \argmin_{\wt \in \chattperm{\permcrl}} \frobnorm{\obs -
  \wt}^2.
\end{align}
The optimization problem~\eqref{EqnDefnCRL} corresponds to a
projection onto the polytope of bi-isotone matrices contained within
the hypercube $[0,1]^\numitems$, along with skew symmetry constraints.
Problems of the form~\eqref{EqnDefnCRL} have been studied in past
work~\cite{bril1984algorithm, robertson1988order, chatterjee2014matrix, kyng2015fast}, and the
estimator $\wtCRL$ is indeed computable in polynomial time.  By
construction, it is agnostic to the values of $(\numcomm, \commset)$.

To provide intuition for the second step of randomization, it serves
to discard ``non-robust'' information from the order computed in Step
1.  Any such information corresponds to noise due to the Bernoulli
sampling process, as opposed to structural information about the
matrix.  If we do not perform this second step---effectively retaining
considerable bias from Step 1---then then isotonic regression
procedure in Step 3 may amplify it, leading to a poorly performing
estimator. To clarify our choice of threshold $T = \sqrt{n} \log
(n)$, the factor $\sqrt{n}$ corresponds to the standard deviation of a
typical win count $\Win_i$ (as a sum of Bernoulli variables), whereas
the $\log n$ serves to control fluctuations in a union bound.

The following theorem provides an upper bound on the adaptivity index
achieved by the CRL estimator.
\begin{theorem}
\label{ThmCRLUpper}
There are universal constants $\UUP$ and $\UUP'$ such that for every
$\wtstar \in \chatterjeeclass$, the CRL estimator $\wtCRL$ has squared
Frobenius norm error
\begin{subequations}
\begin{align}
\label{EqnCRLUpperErr}
\frac{1}{\numitems^2} \frobnorm{\wtCRL - \wtstar}^2 & \leq \UUP
\frac{\numitems - \maxsizefn(\wtstar) + 1}{\numitems^{3/2}} \; (\log
\numitems)^8,
\end{align}
with probability at least $1 - \numitems^{-20}$.  Consequently, its
adaptivity index is upper bounded as
\begin{align}
\label{EqnCRLUpperAdapt} 
\AEsym(\wtCRL) \leq \UUP' \sqrt{\numitems} (\log \numitems)^{8}.
\end{align}
\end{subequations}
\end{theorem}

It is worth noting that equation~\eqref{EqnCRLUpperErr} in yields an
upper bound on the minimax risk of the CRL estimator---namely
\begin{align*}
\sup_{\wtstar \in \chatterjeeclass} \frac{1}{\numitems^2} \Exs [
  \frobnorm{\wtCRL - \wtstar}^2 ] \leq \UUP \frac{ (\log
  \numitems)^8}{\sqrt{\numobs}},
\end{align*}
with this worst-case achieved when $\maxsizefn(\wtstar) = 1$.  Up to
logarithmic factors, this bound matches the best known upper bound on
the minimax rate of polynomial-time estimators~\cite[Theorem
  2]{shah2015stochastically}.


\subsection{A lower bound on adaptivity for polynomial-time algorithms}
\label{SecLower}

By comparing the guarantee~\eqref{EqnCRLUpperAdapt} for the CRL
estimator with the corresponding guarantee~\eqref{EqnRegLseAdaptive}
for the regularized least-squares estimator, we see that (apart from
log factors and constants), their adaptivity indices differ by a
factor of $\sqrt{\numitems}$.  Given this polynomial gap, it is
natural to wonder whether our analysis of the CRL estimator might be
improved, or if not, whether there is another polynomial-time
estimator with a lower adaptivity index than the CRL estimator.  In
this section, we answer \emph{both of these questions in the
  negative,} at least conditionally on a certain well-known conjecture
in average case complexity theory.

More precisely, we prove a lower bound that relies on the average-case
hardness of the planted clique
problem\fullversion{~\cite{jerrum1992large,kuvcera1995expected}}.  The
use of this conjecture as a hardness assumption is widespread in the
literature~\cite{juels2000hiding, alon2007testing_short,
  dughmi2014hardness}, and there is now substantial evidence in the
literature supporting the conjecture~\cite{jerrum1992large,
  feige2003probable, meka2015sum, deshpande2015improved}.  It has
also been used as a tool in proving hardness results for sparse PCA
and related matrix recovery problems~\cite{rigollet, ma2015computational}.

In informal terms, the planted clique conjecture asserts that it is
hard to detect the presence of a planted clique in an
Erd\H{o}s-R\'enyi random graph.  In order to state it more precisely,
let $\graph(\numitems, \planted)$ be a random graph on $\numitems$
vertices constructed in one of the following two ways:
\begin{itemize}
\item[$H_0$:] Every edge is included in $\graph(\numitems, \planted)$
  independently with probability $\half$.
\item[$H_1$:] Every edge is included in $\graph(\numitems, \planted)$
  independently with probability $\half$. In addition, a set of
  $\planted$ vertices is chosen uniformly at random and all edges with
  both endpoints in the chosen set are added to $\graph$.
\end{itemize}
The planted clique conjecture then asserts that when \mbox{$\planted =
  o(\sqrt{\numitems})$}, then there is no polynomial-time algorithm
that can correctly distinguish between $H_0$ and $H_1$ with an error
probability that is strictly bounded below $1/2$.

\noindent Using this conjectured hardness as a building block, we have
the following result:

\begin{theorem} 
\label{ThmPolyLower}
Suppose that the planted clique conjecture holds. Then there is a
universal constant $\ULOW > 0$ such that for any polynomial-time
computable estimator $\wthat$, its adaptivity index is lower bounded
as
\begin{align*}
\AEsym(\wthat) \geq \ULOW \sqrt{\numitems} (\log \numitems)^{-3}.
\end{align*}
\end{theorem}

Together, the upper and lower bounds of Theorems~\ref{ThmCRLUpper}
and~\ref{ThmPolyLower} imply that the estimator $\wtCRL$ achieves the
optimal adaptivity index (up to logarithmic factors) among all
computationally efficient estimators.


\subsection{Negative results for the least squares estimator}
\label{SecLSEbad}

In this section, we study the adaptivity of the (unregularized) least
squares estimator given by
\begin{align}
\label{EqnDefnLSE}
\wtlse \in \argmin \limits_{\wt \in \chatterjeeclass} \frobnorm{\obs - \wt}^2.
\end{align}
Least squares estimators of this type are known to possess very good
adaptivity in various other problems of shape-constrained estimation;
for instance, see the papers~\cite{cator2011adaptivity,
  chatterjee2013risk, chatterjee2015adaptive,
  chatterjee2015biisotonic, bellec2016adaptive} and references therein
for various examples of such phenomena.  From our own past
work~\cite{shah2015stochastically}, the estimator~\eqref{EqnDefnLSE}
is known to be minimax optimal for estimating SST matrices.

Given this context, the following theorem provides a surprising
result---namely, that the least-squares estimator~\eqref{EqnDefnLSE}
has remarkably poor adaptivity:
\begin{theorem}
\label{thm:break_lse}
There is a universal constant $\ULOW > 0$ such that the adaptivity
index of the least squares estimate~\eqref{EqnDefnLSE} is lower
bounded as
\begin{align}
\label{EqnPoor}
\AEsym(\wtlse) \geq \ULOW \; \numitems \; (\log \numitems)^{-2}.
\end{align}
\end{theorem}
\noindent In order to understand why the lower bound~\eqref{EqnPoor}
is very strong, consider the trivial estimator $\Mbar$ that simply
\emph{ignores the data}, and returns the constant matrix $\Mbar =
\half \ones \ones^T$.  It can be verified that we have
\begin{align*}
\frobnorm{\wtstar - \half \ones \ones^T}^2 & \leq 3 \numitems
(\numitems - \maxsizefn(\wtstar) + 1)
\end{align*}
for every $\wtstar \in \chatterjeeclass$. Thus, for this trival
estimator $\Mbar$, we have $\AEsym(\Mbar) \leq \UUP \numitems$.
Comparing to the lower bound~\eqref{EqnPoor}, we see that apart from
logarithmic factors, the adaptivity of the least squares estimator is
no better than that of the trivial estimator $\Mbar$.

\section{Proofs}
\label{SecProofs}

In this section, we present the proofs of our results.  We note in
passing that our proofs additionally lead to some auxiliary results
that may be of independent interest. These auxiliary results pertain
to the problem of bivariate isotonic regression---that is, estimating
$\wtstar$ when the underlying permutation is known---an important
problem in the field of shape-constrained
estimation~\cite{robertson1988order, tibshirani2011nearly,
  chatterjee2014matrix}. Prior works restrict attention to the
expected error and assume that the underlying permutation is correctly
specified; our results provide exponential tail bounds and also
address settings when the permutation is misspecified.

A few comments about assumptions and notation are in order.  In all of
our proofs, so as to avoid degeneracies, we assume that the number of
items $\numitems$ is greater than a universal constant.  (The cases
when $\numitems$ is smaller than some universal constant all follow by
adjusting the pre-factors in front of our results suitably.)  For any
matrix $\wt$, we use $\maxsizefn(\wt)$ to denote the size of the
largest indifference set in $\wt$, and we define $\commstar =
\maxsizefn(\wtstar)$. The notation $\plaincon, \plaincon_1, \UUP,
\ULOW$ etc. all denote positive universal constants. For any two
square matrices $A$ and $B$ of the same size, we let $\tracer{A}{B} =
\trace(A^T B)$ denote their trace inner product. For an $(\numitems
\times \numitems)$ matrix $\wt$ and any permutation $\perm$ on
$\numitems$ items, we let $\perm(\wt)$ denote an $(\numitems \times
\numitems)$ matrix obtained by permuting the rows and columns of $\wt$
by $\perm$.  For a given class $\mathbb{C}$ of matrices, metric $\rho$
and tolerance $\epsilon > 0$, we use
$\covnum(\epsilon,\mathbb{C},\rho)$ to denote the $\epsilon$ covering
number of the class $\mathbb{C}$ in the metric $\rho$. The metric
entropy is given by the logarithm of the covering number---namely
$\metent(\epsilon,\mathbb{C},\rho)$.

It is also convenient to introduce a linearized form of the
observation model~\eqref{EqnDefnBernModel}.  Observe that we can write the observation matrix
$\obs$ in a linearized fashion as
\begin{subequations}
\begin{align}
\obs = \wtstar + \noise,
\label{EqnLinearized}
\end{align}
where $\noise \in [-1,1]^{\numitem \times \numitem}$ is a random
matrix with independent zero-mean entries for every $i > j$, and and
$\noise_{ji} = - \noise_{ij}$ for every $i < j$.  For $i > j$, its
entries follow the distribution
\begin{align}
\label{EqnDefnNoise}
\noise_{ij} & \sim \begin{cases} 1 - \Mstar_{ij} & \mbox{with
    probability $\Mstar_{ij}$} \\
-\Mstar_{ij} & \mbox{with probability $1 - \Mstar_{ij}$}.
\end{cases}
\end{align}
\end{subequations}
In summary, all entries of the matrix $\noise$ above the main diagonal
are independent, zero-mean, and uniformly bounded by $1$ in absolute
value.  This fact plays an important role in several parts of our
proofs. \\



\subsection{A general upper bound on regularized $M$-estimators}

In this section, we prove a general upper bound that applies to a
relatively broad class of regularized $M$-estimators for SST matrices.
Given a matrix $\obs$ generated from the model~\eqref{EqnLinearized},
consider an estimator of the form
\begin{align}
\label{EqnMest}
\wthatgenup \in \argmin \limits_{\wt \in \classgenup} \Big \{
\frobnorm{ \obs - \wt }^2 + \reggenup(\wt) \Big \}.
\end{align}
Here $\reggenup: [0,1]^{\numitems \times \numitems} \rightarrow
\reals_+$ is a regularization function to be specified by the user,
and $\classgenup$ is some subset of the class $\chatterjeeclass$ of
SST matrices.  Our goal is to derive a high-probability bound on the
Frobenius norm error $\frobnorm{\wthatgenup - \Mstar}$.  As is
well-known from theory on
$M$-estimators~\cite{van2000empirical,Bar05,Kolt06}, doing so requires
studying the empirical process in a localized sense.

In order to do so, it is convenient to consider sets of the form
\begin{align*}
\chattDiff(\wtstar, \ballgenup,\bgenup, \mathbb{C}) \defn \{
\alpha(\wt - \wtstar) \mid \wt \in \classgenup, \alpha \in [0,1],~
\bgenup\frobnorm{ \alpha(\wt - \wtstar)} \leq \bgenup \ballgenup \},
\end{align*}
where $t \in [0,n]$, and $\bgenup \in \{0,1\}$. The binary flag
$\bgenup$ controls whether or not the set is localized around
$\wtstar$, and the radius $t$ controls the extent to which the set is
localized.

In the analysis to follow, we assume that for each $\epsilon \geq
\numitems^{-8}$, the $\epsilon$-metric entropy of $\chattDiff(\wtstar,
\ballgenup,\bgenup, \mathbb{C})$ satisfies an upper bound of the form
\begin{subequations}
\begin{align}
\label{EqnGenupMetent}
\metent(\epsilon, \chattDiff(\wtstar, \ballgenup,\bgenup, \mathbb{C})
, \frobnorm{\cdot} ) \leq \frac{t^{2 \bgenup}
  (\ggenup(\wtstar))^2}{\epsilon^2} + (\hgenup(\wtstar))^2,
\end{align}
where $\ggenup: \mathbb{R}^{\numitems \times \numitems} \mapsto
\mathbb{R}_+$ and $\hgenup: \mathbb{R}^{\numitems \times \numitems}
\mapsto \mathbb{R}_+$ are some functions.  In the sequel, we provide
concrete examples of sets $\mathbb{C}$ and functions $(\ggenup,
\hgenup)$ for which a bound of this form holds.

Given $(\ggenup, \hgenup, \reggenup)$, we can then define a critical radius $\delcrit \geq 0$ as
\begin{align}
\label{EqnCritRad}
\delcrit^2 = \plaincon \big( ( \ggenup(\wtstar) \log
\numitems)^{1+\bgenup} + (\hgenup(\wtstar))^2 + \reggenup(\wtstar) +
\numitems^{-7} \big),
\end{align}
\end{subequations}
where $\plaincon > 0$ is a universal constant.  The following
result guarantees that the Frobenius norm can be controlled
by the square of this critical radius:

\begin{lemma}
\label{LemGeneralUpper}
For any set $\mathbb{C}$ satisfying the metric entropy
bound~\eqref{EqnGenupMetent}, the Frobenius norm of the
$M$-estimator~\eqref{EqnMest} can be controlled as
\begin{align}
\mprob \Big[ \frobnorm{\wthat - \wtstar}^2 > \tailgenup \delcrit^2
  \Big] & \leq e^{- \tailgenup \delcrit^2} \qquad \mbox{for all
  $\tailgenup \geq 1$},
\end{align}
where $\delcrit$ is the critical radius~\eqref{EqnCritRad}.
\end{lemma}
\noindent The significance of this claim is that it reduces the
problem of controlling the error in the $M$-estimator to bounding the
metric entropy (as in equation~\eqref{EqnGenupMetent}), and then
computing the critical radius~\eqref{EqnCritRad}.  The remainder of
this section is devoted to the proof of this claim.


\subsubsection{Proof of Lemma~\ref{LemGeneralUpper}}

Define the difference $\DelHat = \wthatgenup - \wtstar$ between
$\wtstar$ and the optimal solution $\wthatgenup$ to the constrained
least-squares problem.  Since $\wthatgenup$ is optimal and $\Mstar$ is
feasible, we have
\begin{align*}
\frobnorm{\obs - \wthatgenup}^2 + \reggenup(\wthatgenup) \leq
\frobnorm{\obs - \wtstar}^2 + \reggenup(\wtstar).
\end{align*}
Following some algebra, and using the assumed non-negativity condition
$\reggenup(\cdot) \geq 0$, we arrive at the basic inequality
\begin{align*}
\frac{1}{2}
\frobnorm{\DelHat}^2 \leq \tracer{\DelHat}{\Wmat} + \reggenup(\wtstar),
\end{align*}
where $\Wmat \in [0,1]^{\numitems \times \numitems}$ is the noise
matrix in the linearized observation model~\eqref{EqnLinearized}, and
$\tracer{\DelHat}{\Wmat}$ denotes the trace inner product between
$\DelHat$ and $\Wmat$.

Now define the supremum $Z(t) \defn \sup \limits_{D \in
  \chattDiff(\wtstar, \ballgenup,\bgenup, \mathbb{C}) }
\tracer{D}{W}$.  With this definition, we find that the error matrix
$\DelHat$ satisfies the inequality
\begin{align}
\label{EqnBasic}
\frac{1}{2} \frobnorm{\DelHat}^2 & \leq \tracer{\DelHat}{\Wmat} +
\reggenup(\wtstar) \; \leq \; Z \big( \frobnorm{\DelHat} \big) +
\reggenup(\wtstar).
\end{align}
Thus, in order to obtain a high probability bound, we need to
understand the behavior of the random quantity $Z(t)$.

By definition, the set $\chattDiff(\wtstar, \ballgenup,\bgenup,
\mathbb{C})$ is star-shaped, meaning that $\alpha \chattDiffmx \in
\chattDiff(\wtstar, \ballgenup,\bgenup, \mathbb{C})$ for every $\alpha
\in [0,1]$ and every $\chattDiffmx \in \chattDiff(\wtstar,
\ballgenup,\bgenup, \mathbb{C})$. Using this star-shaped property, it
is straightforward to verify that $Z(t)$ grows at most linearly with
$t$, ensuring that there is a non-empty set of scalars $t > 0$
satisfying the critical inequality:
\begin{align}
\label{EqnCritical}
\Exs[Z(t)] + \reggenup(\wtstar) & \leq \frac{t^2}{2}.
\end{align}
Our interest is in an upper bound on the smallest (strictly) positive
solution $\delcrit$ to the critical inequality \eqref{EqnCritical}.
Moreover, our goal is to show that for every $t \geq \delcrit$, we
have $\frobnorm{\DelHat}^2 \leq c t \delcrit$ with probability at
least $1 - c_1 e^{-c_2 t \delcrit}$.

Define the ``bad'' event
\begin{align}
\label{EqnDefnBadevent}
\AuxEvent & \defn \big \{ \exists \Delta \in
\chattDiff(\wtstar,\ballgenup) \mid \frobnorm{\Delta} \geq
\sqrt{\ballgenup \delcrit} \quad \mbox{and} \quad
\tracer{\Delta}{\Wmat} + \reggenup(\wtstar) \geq 2 \frobnorm{\Delta}
\sqrt{\ballgenup \delcrit} \big \}.
\end{align}
Using the star-shaped property of $\chattDiff(\wtstar,
\ballgenup,\bgenup, \mathbb{C})$ and the fact that $\reggenup(\cdot)
\geq 0$, it follows by a rescaling argument that
\begin{align*}
\mprob[\AuxEvent] \leq \mprob[Z(\delcrit) + \reggenup(\wtstar) \geq 2
  \delcrit \sqrt{t \delcrit}] \qquad \mbox{for all $t \geq \delcrit$.}
\end{align*}
The entries of $\Wmat$ lie in $[-1,1]$, have a mean of zero, are
i.i.d. on and above the diagonal, and satisfy skew-symmetry.
Moreover, the function $\Wmat \mapsto Z(u)$ is convex and Lipschitz
with parameter $u$.  Consequently, by Ledoux's concentration
theorem~\cite[Theorem 5.9]{Ledoux01}, we have
\begin{align*}
\mprob \big[ Z(\delcrit) \geq \Exs[Z(\delcrit)] + \sqrt{t \delcrit}
  \delcrit \big] & \leq e^{-c_1 t \delcrit} \qquad \mbox{for all $t
  \geq \delcrit$,}
\end{align*}
for some universal constant $c_1$.  By the definition of $\delcrit$,
we have $\Exs[Z(\delcrit)] + \reggenup(\wtstar) \leq \delcrit^2 \leq
\delcrit \sqrt{t \delcrit}$ for any $t \geq \delcrit$, and
consequently
\begin{align*}
\mprob[\AuxEvent] & \leq \mprob[Z(\delcrit)+ \reggenup(\wtstar) \geq 2
  \delcrit \sqrt{t \delcrit} \big] \; \leq \; e^{-c_1 t \delcrit}
\quad \mbox{for all $t \geq \delcrit$.}
\end{align*}
Consequently, either $\frobnorm{\DelHat} \leq \sqrt{t \delcrit}$, or
we have $\frobnorm{\DelHat} > \sqrt{t \delcrit}$. In the latter case,
conditioning on the complement $\AuxEvent^c$, the basic
inequality~\eqref{EqnBasic} implies that $\frac{1}{2}
\frobnorm{\DelHat}^2 \leq 2 \frobnorm{\DelHat} \sqrt{t \delcrit}$.
Putting together the pieces yields that
\begin{align*}
\frobnorm{\DelHat} \leq 4 \sqrt{t \delcrit},
\end{align*}
with probability at least $1- e^{-c_1 t \delcrit}$ for every $t \geq
\delcrit$.  Substituting $\tailgenup = \frac{\ballgenup}{\delcrit}$,
we get
\begin{align}
\label{EqWithDelCritHP}
\mprob \Big( \frobnorm{\DelHat}^2 > c_2 \tailgenup \delcrit^2 \Big) \leq e^{-c_1 \tailgenup \delcrit^2},
\end{align}
for every $\tailgenup \geq 1$.

In order to determine a feasible $\delcrit$ satisfying the critical
inequality~\eqref{EqnCritical}, we need to bound the expectation
$\Exs[Z(\delcrit)]$.  To this end, we introduce an auxiliary lemma:
\begin{lemma}
\label{LemCritical}
There is a universal constant $\plaincon$ such that for any set
$\mathbb{C}$ satisfying the metric entropy
bound~\eqref{EqnGenupMetent}, we have
\begin{align}
\label{EqnCriticalBound}
\Exs[Z(t)] & \leq \plaincon \, \Big \{ \ballgenup^{\bgenup}
\ggenup(\wtstar) \log \numitems + t \, \hgenup(\wtstar) +
\numitems^{-7} \Big \} \qquad \mbox{for all $t 
\geq 0$.}
\end{align}
\end{lemma}
\noindent See Section~\ref{SecProofLemCritical} for the proof of this
claim.

\noindent Using Lemma~\ref{LemCritical}, we see that the
critical inequality~\eqref{EqnCritical} is satisfied for 
\begin{align*}
\delcrit = \plaincon_0 \Big\{ \big(\ggenup(\wtstar) \log \numitems
\big)^{\half(\bgenup + 1)} + \, \hgenup(\wtstar) +
\sqrt{\reggenup(\wtstar)} + \numitems^{-\frac{7}{2}} \Big\},
\end{align*} 
for a positive universal constant $\plaincon_0$.  With this choice, our claim follows from
the tail bound~\eqref{EqWithDelCritHP}, absorbing the constants $\plaincon_1$ and $\plaincon_2$ into $\plaincon_0$. \\

\noindent It remains to prove Lemma~\ref{LemCritical}.


\subsubsection{Proof of Lemma~\ref{LemCritical}}
\label{SecProofLemCritical}

By the truncated form of Dudley's entropy inequality, we have
\begin{align}
\Exs[ Z(t) ] & \leq \plaincon\; \inf_{\delta \in [0,\numitems]} \Big
\{ \numitems \delta + \int_{\frac{\delta}{2}}^{t} \sqrt{
  \metent(\epsilon,\chattDiff(\wtstar, \ballgenup,\bgenup, \mathbb{C})
  ,\frobnorm{.})}  d\epsilon \Big \} \notag \\
\label{EqnEarly}
& \leq \plaincon \; \Big \{ 2\numitems^{-7} +
\int_{\numitems^{-8}}^{t} \sqrt{ \metent(\epsilon,\chattDiff(\wtstar,
  \ballgenup,\bgenup, \mathbb{C}) ,\frobnorm{.})} d\epsilon \Big \},
\end{align}
where the second step follows by setting $\delta = 2 \numitems^{-8}$.
Combining our assumed upper bound~\eqref{EqnGenupMetent} on the metric
entropy with the earlier inequality~\eqref{EqnEarly} yields
\begin{align*}
\Exs [Z(t) ] & \leq \plaincon \big \{ 2 \numitems^{-7} +
\ballgenup^\bgenup \ggenup(\wtstar) \log (\numitems \ballgenup) +
\ballgenup \hgenup(\wtstar) \big \} \; \leq \; 2 \plaincon \big \{ 2
\numitems^{-7} + \ballgenup^\bgenup \ggenup(\wtstar) \log \numitems +
\ballgenup \hgenup(\wtstar) \big \},
\end{align*}
where the final step uses the upper bound $t \leq \numitems$. We have
thus established the claimed bound~\eqref{EqnCriticalBound}.



\subsection{Proof of Proposition~\ref{PropORA}}
\label{SecProofPropORA}

We are now equipped to prove bounds on the risk incurred by the oracle
estimator from equation~\eqref{EqnOracleRisk}.


\subsubsection{Upper bound}

Let $\commstar = \Lnorm{\commset}{\infty}$ denote the size of the
largest indifference set in $\wtstar$, and recall that the oracle
estimator knows the value of $\commstar$.  For our upper bound, we use
Lemma~\ref{LemGeneralUpper} from the previous section with
\begin{align*}
\classgenup = \chatterjeeclass(\commstar), \qquad \reggenup(\wt) = 0,
\quad \mbox{and} \quad \bgenup = 0.
\end{align*}
With these choices, the estimator~\eqref{EqnMest} for which
Lemma~\ref{LemGeneralUpper} provides guarantees is equivalent to the
oracle estimator~\eqref{EqnOracleRisk}. We then have
\begin{align*}
\chattDiff(\wtstar, \ballgenup, \chatterjeeclass(\commstar) ) = \Big
\{ \alpha(\wt - \wtstar) \mid \wt \in \classgenup, \alpha \in [0,1]
\Big \}.
\end{align*}
In order to apply the result of Lemma~\ref{LemGeneralUpper}, we need
to compute the metric entropy of the set $\chattDiff$.  For ease of
exposition, we further define the set
\begin{align*}
\chattstar(k) \defn \{ \alpha \wt \mid \wt \in \chatterjeeclass(k), \,
\alpha \in [0,1] \}.
\end{align*}
Since $\wtstar \in \chatterjeeclass(\commstar)$, the metric entropy of
$\chattDiff$ is at most twice the metric entropy of
$\chattstar(\commstar)$.  The following lemma provides an upper bound
on the metric entropy of the set $\chattstar(k)$:
\begin{lemma}
\label{LemChattMetent_adaptive}
For every $\epsilon >0$ and every integer $k \in [\numitems]$, the metric entropy
is bounded as
\begin{align}
\label{EqnGobiManch}
\metent(\epsilon, \chattstar(k), \frobnorm{.}) & \leq \plaincon
\frac{(\numitems - k + 1)^2}{\epsilon^2} \Big(\log
\frac{\numitems}{\epsilon} \Big)^2 + \plaincon (\numitems - k + 1)
\log \numitems,
\end{align}
where $\plaincon > 0$ is a universal constant.
\end{lemma}

With this lemma, we are now equipped to prove the upper bound in
Proposition~\ref{PropORA}.  The bound~\eqref{EqnGobiManch} implies
that
\begin{align*}
\metent(\epsilon, \chattDiff(\wtstar, \ballgenup,
\chatterjeeclass(\commstar)), \frobnorm{.}) & \leq \plaincon'
\frac{(\numitems - \commstar + 1)^2}{\epsilon^2} (\log \numitems )^2 +
\plaincon' (\numitems - \commstar + 1) \log \numitems,
\end{align*}
for all $\epsilon \geq \numitems^{-8}$.  Consequently, a bound of the
form~\eqref{EqnGenupMetent} holds with $\ggenup(\wtstar) =  \sqrt{\plaincon'} (\numitems
- \commstar + 1) \log \numitems$ and $\hgenup(\wtstar) =
\sqrt{\plaincon' (\numitems - \commstar + 1) \log \numitems}$.  Applying
Lemma~\ref{LemGeneralUpper} with $\tailgenup = 1$ yields
\begin{align*}
\mprob \big( \frobnorm{ \wttil(\numcomm,\commset) - \wtstar}^2 >
\plaincon (\numitems - \commstar + 1) (\log \numitems)^2 \big) \leq
e^{- (\numitems -\commstar + 1) (\log \numitems)^2},
\end{align*}
where $\plaincon > 0$ is a universal constant. Integrating this tail
bound (and using the fact that the Frobenius norm is bounded as
\mbox{$\frobnorm{\wttil(\numcomm,\commset) - \wtstar} \leq
  \numitems$)} gives the claimed result.


\subsubsection{Lower bound}

We now turn to proving the lower bound in Proposition~\ref{PropORA}.
By re-ordering as necesseary, we may assume without loss of generality
that $\commsize_1 \geq \cdots \geq \commsize_\numcomm$, so that
$\commmax = \commsize_1$. The proof relies on the following technical
preliminary that establishes a lower bound on the minimax rates of
estimation when there are two indifference sets.
\begin{lemma}
\label{lem:twocomm}
If there are $\numcomm = 2$ indifference sets (say, of sizes
$\commsize_1 \geq \commsize_2$), then any estimator $\wthat$ has error
lower bounded as
\begin{align}
\label{EqnRoti}
\sup_{\wtstar \in \chatterjeeclass(2, (\commsize_1,\commsize_2))}
\frac{1}{\numitems^2} \Exs [ \frobnorm{\wthat - \wtstar}^2 ] \geq
\ULOW \frac{\numitems - \commsize_1}{\numitems^2}.
\end{align}
\end{lemma}
\noindent See Section~\ref{SecProofLemTwocomm} for the proof of this
claim.

Let us now complete the proof of the lower bound in
Proposition~\ref{PropORA}. We split the analysis into two cases
depending on the size of the largest indifference set.

~\\\underline{Case I:} First, suppose that $\commsize_1 >
\frac{\numitems}{3}$.  We then observe that $\chatterjeeclass(2,(k_1,
n-k_1))$ is a subset of $\chatterjeeclass(\commset)$: indeed, every
matrix in $\chatterjeeclass(2,(k_1, n-k_1))$ can be seen as a matrix
in $\chatterjeeclass(\commset)$ which has identical values in entries
corresponding to all items not in the largest indifference set.  Since
the induced set $\chatterjeeclass(2,(k_1, n-k_1))$ is a subset of
$\chatterjeeclass(\commset)$, the lower bound for estimating a matrix
in $\chatterjeeclass(\commset)$ is at least as large as the lower
bound for estimating a matrix in the class $\chatterjeeclass(2,(k_1,
n-k_1))$.  Now applying Lemma~\ref{lem:twocomm} to the set
$\chatterjeeclass(2,(k_1, n-k_1))$ yields a lower bound of $\ULOW
\frac{\min \{\numitems - \commsize_1,
  \commsize_1\}}{\numitems^2}$. Since $\commsize_1 >
\frac{\numitems}{3}$, we have $\commsize_1 \geq \frac{\numitems -
  \commsize_1}{2}$. As a result, we get a lower bound of
$\frac{\ULOW}{2} \frac{\numitems - \commsize_1}{\numitems^2}$.

~\\\underline{Case II:} Alternatively, suppose that $\commsize_1 \leq
\frac{\numitems}{3}$.  In this case, we claim that there exists a
value $u \in [n/3, 2n/3]$ such that $\chatterjeeclass(2,
(u,\numitems-u))$ is a subset of the set $\chatterjeeclass(\commset)$
with $\commsize_1 \leq \frac{\numitems}{3}.$ Observe that for any
collection of sets with sizes $\commset$ with $\commsize_1 \leq
\frac{\numitems}{3}$, there is a grouping of sets into two groups,
both of size between $\numitems/3$ and $2\numitems/3$. This is true
since the largest set is of size at most $\numitems/3$.  Denoting the
size of either of these groups as $u$, we have established our earlier
claim.

As in the previous case, we can now apply Lemma~\ref{lem:twocomm} to
the subset $\chatterjeeclass(2, (u,\numitems-u))$ to obtain a lower
bound of $\ULOW \frac{1}{3\numitems} \geq \ULOW \frac{ \numitems -
  \commsize_1}{3\numitems^2}$.


\subsubsection{Proof of Lemma~\ref{LemChattMetent_adaptive}}

In order to upper bound the metric entropy of $\chattstar(k)$, we
first separate out the contributions of the permutation and the
bivariate monotonicity conditions.  Let
$\chattperm[\widetilde]{\permid}(k)$ denote the subset of matrices in
$\chattstar(k)$ that are faithful to the identity permutation.  With
this notation, the $\epsilon$-metric entropy of
$\chattstar(\commstar)$ is upper bounded by the sum of two parts: 
\begin{enumerate}[leftmargin = *]
\item[(a)] the $\epsilon$-metric entropy of the set
  $\chattperm[\widetilde]{\permid}(k)$; and 
\item[(b)] the logarithm of the number of distinct permutations of the
  $\numitems$ items. 
\end{enumerate}
Due to the presence of an indifference set of size at least $k$, the
quantity in (b) is upper bounded by $\log (\frac{\numitems!}{k!}) \leq
(\numitems - k) \log \numitems$.

We now upper bound the $\epsilon$-metric entropy of the set
$\chattperm[\widetilde]{\permid}(k)$. We do so by partitioning the
$\numitems^2$ positions in the matrix, computing the $\epsilon$-metric
entropy of each partition separately, and then adding up these metric
entropies. More precisely, letting $S_k \subseteq [\numitems]$ denote some set of $k$ items that belong to the same indifference set, let us partition the entries of each matrix
into four sub-matrices as follows: 
\begin{enumerate}[leftmargin = *]
\item[(i)] The $(k \times k)$ sub-matrix comprising entries $(i,j)$
  where both $i \in S_k$ and $j \in S_k$;
\item[(ii)] the $(k \times (\numitems - k))$ sub-matrix comprising
  entries $(i,j)$ where $i \in S_k$ and
  $j \in [\numitems]\backslash S_k$;
\item[(iii)] $((\numitems - k ) \times k)$ sub-matrix comprising
  entries $(i,j)$ where $i \in [\numitems]\backslash S_k$ and $j \in S_k$; and
\item[(iv)] the $((\numitems - k ) \times (\numitems - k))$ sub-matrix
  comprising entries $(i,j)$ where both $i \in [\numitems]\backslash S_k$ and $j \in [\numitems]\backslash S_k$.
\end{enumerate}
By construction, the metric entropy of
$\chattperm[\widetilde]{\permid}(k)$ is at most the sum of the metric
entropies of these sub-matrices.

The set of sub-matrices in (i) comprises only constant matrices, and
hence its metric entropy is at most $\log
\frac{\numitems}{\epsilon}$. Any sub-matrix from set (ii) has
constant-valued columns, and so the metric entropy of this set is
upper bounded by $(\numitems - k) \log \frac{\numitems}{\epsilon}$. An
identical bound holds for the set of sub-matrices in (iii). Finally,
the set of sub-matrices in (iv) are all contained in the set of all
$((\numitems - k ) \times (\numitems - k))$ SST matrices. The metric
entropy of the SST class is analyzed in Theorem 1 of our past
work~\cite{shah2015stochastically}, where we showed that the metric
entropy of this set is at most $2\left(\frac{\numitems -
  k}{\epsilon}\right)^2 \big(\log \frac{\numitems - k}{\epsilon}\big)^2 +
(\numitems - k) \log \numitems$. Summing up each of these metric
entropies, some algebraic manipulations yield the claimed result.

\subsubsection{Proof of Lemma~\ref{lem:twocomm}}
\label{SecProofLemTwocomm}

For the first part of the proof, we assume $\commsize_2$ is greater
than a universal constant.  (See the analysis of Case 2 below for how
to handle small values of $\commsize_2$.)  Under this condition, the
Gilbert-Varshamov
bound~\cite{gilbert1952comparison,varshamov1957estimate} guarantees
the existence of a binary code $\CODE$ of length $\commsize_2$,
minimum Hamming distance $\plaincon_0 \commsize_2$, and number of code
words $\card(\CODE) = \packnum = 2^{\plaincon \commsize_2}$.  (As
usual, the quantities $\plaincon$ and $\plaincon_0$ are positive
numerical constants.)

We now construct a set of $\packnum$ matrices contained within the set
$\chatterjeeclass(2, (\commsize_1,\commsize_2))$, whose constituents
have a one-to-one correspondence with the $\packnum$ codewords of the
binary code constructed above.  Let items \mbox{$S = \{1, \ldots,
  \commsize_1 \}$} correspond to the first indifference set, so that
the complementary set \mbox{$S^c \defn \{\commsize_1 + 1, \ldots,
  \numitems \}$} indexes the second indifference set.

Fix some $\delta \in (0,\frac{1}{3}]$, whose precise value is to be
specified later. Define the base matrix $\MZERO$ with entries
\begin{align*}
\MZEROSUB{ij} & = \begin{cases} 
                  \half & \mbox{if $i,j \in S$ or $i, j \in S^c$} \\
                  \half + \delta & \mbox{if $i \in S$ and $j \in S^c$} \\
                  \half - \delta & \mbox{if $i \in S^c$ and $j \in S$}.
\end{cases}
\end{align*}
For any other codeword $\cword \in \CODE$, the matrix $M(\cword)$ is
defined by starting with the base matrix $\MZERO$, and then swapping
row/column $i$ with row/column $(\commsize_1 + i)$ if and only if
$\cword_i = 1$.  For instance, if the codeword is $\cword =
[1~1~0~\cdots~0]$, then the new ordering in the matrix $M(\cword)$ is
given by
$(\commsize_1+1),(\commsize_1+2),3,\ldots,\commsize_1,1,2,(\commsize_1+3),\ldots,\numitems$,
which is obtained by swapping the first two items of the two
indifference sets.

We have thus constructed a set of $\packnum$ matrices that are
contained within the set $\chatterjeeclass(2,
(\commsize_1,\commsize_2))$. We now evaluate certain properties of
these matrices which will allow us prove the claimed lower
bound. Consider any two matrices $\wt_1$ and $\wt_2$ in this
set. Since any two codewords in our binary code have a Hamming
distance at least $\plaincon_0 \commsize_2$, we have from the
aforementioned construction:
\begin{align*}
\plaincon_1 \commsize_2 \numitems \delta^2  \leq \frobnorm{\wt_1 - \wt_2}^2 \leq 2 \commsize_2 \numitems \delta^2,
\end{align*}
for a constant $\plaincon_1 \in (0,1)$.

Let $\mprob_{M_1}$ and $\mprob_{M_2}$ correspond to the distributions
of the random matrix $Y$ based on Bernoulli sampling~\eqref{EqnDefnBernModel} from the matrices
$M_1$ and $M_2$, respectively.  Since $\delta \in (0,\frac{1}{3}]$,
  all entries of the matrices $M_1$ and $M_2$ lie in the interval
  $[1/3, 2/3]$.  Under this boundedness condition, the KL divergence
  may be sandwiched by the Frobenius norm up to constant factors.  Applying
  this result in the current setting yields
\begin{align*}
\plaincon_2 \commsize_2 \numitems \delta^2 \leq
\kl{\mprob_{\wt_1}}{\mprob_{\wt_2}} \leq \plaincon_3 \commsize_2
\numitems \delta^2,
\end{align*}
again for positive universal constants $\plaincon_2$ and
$\plaincon_3$.  An application of Fano's inequality to this set gives
that the error incurred by any estimator $\wthat$ is lower bounded as
\begin{align}
\label{EqnFano}
\sup_{\wtstar \in \chatterjeeclass(2, (\commsize_1,\commsize_2))} \Exs
    [ \frobnorm{\wthat - \wtstar}^2 ] \geq \frac{\plaincon_1
      \commsize_2 \numitems \delta^2}{2} \left( 1 - \frac{\plaincon_3
      \commsize_2 \numitems \delta^2 + \log 2}{\plaincon \commsize_2}
    \right).
\end{align}

\noindent From this point, we split the remainder of the analysis into
two cases.

\paragraph{Case 1:}  First suppose that $\commsize_2$ is larger than
some suitably large (but still universal) constant.  In this case, we
may set $\delta^2 = \frac{\plaincon''}{\numitems }$ for a small enough
universal constant $\plaincon''$, and the Fano bound~\eqref{EqnFano}
then implies that
\begin{align*}
\sup_{\wtstar \in \chatterjeeclass(2, (\commsize_1,\commsize_2))} \Exs
    [ \frobnorm{\wthat - \wtstar}^2 ] & 
 \geq \plaincon' \commsize_2,
\end{align*}
for some universal constant $\plaincon' > 0$.  Since $\commsize_2 =
\numitems - \commsize_1$, this completes the proof the claimed lower
bound~\eqref{EqnRoti} in this case.

\paragraph{Case 2:}  Otherwise,
the parameter $\commsize_2$ is smaller than the universal constant in
the above part of the proof. In this case, the claimed lower
bound~\eqref{EqnRoti} on $\Exs [\frobnorm{\wthat - \wtstar}^2]$ is
just a constant, and we can handle this case with a different
argument.  In particular, suppose that the estimator is given
partition forming the two indifference sets, and only needs to
estimate the parameter $\delta$. For this purpose, the sufficient
statistics of the observation matrix $Y$ are those entries of the
observation matrix that correspond to matches between two items of
different indifference sets; note that there are $\commsize_1
\commsize_2$ such entries in total. From standard bounds on estimation of a single Bernoulli
probability, any estimator $\hat{\delta}$ of $\delta$ must have
mean-squared error lower bounded as $\Exs[ (\delta - \hat{\delta})^2]
\geq \frac{\plaincon}{\commsize_1 \commsize_2}$. Finally, observe that
the error in estimating the matrix $\wtstar$ in the squared Frobenius
norm is at least $2 \commsize_1 \commsize_2$ times the error in
estimating the parameter $\delta$.  We have thus established the
claimed lower bound of a constant.


\subsection{Proof of Theorem~\ref{ThmRegLse}}

We now prove the upper bound~\eqref{EqnRegLseUpper} for the
regularized least squares estimator~\eqref{EqnDefnRegLse}.  Note that
it has the equivalent representation
\begin{align}
\label{EqnDefnRegRewrite}
\wtreg \in \argmin \limits_{\wt \in \chatterjeeclass} \Big \{
\frobnorm{\obs - \wt}^2 + \reg{\maxsizefn(\wt)} \Big \}.
\end{align}
Defining $\commstar \defn \commmax(\wtstar)$, it is also convenient to
consider the family of estimators
\begin{align}
\label{EqnDefnRegWeaker}
\wtcomm \in \argmin \limits_{\wt \in \chatterjeeclass(\commsize) \cup
  \chatterjeeclass(\commstar)} \Big \{ \frobnorm{\obs - \wt}^2 +
\reg{\maxsizefn(\wt)} \Big \},
\end{align}
where $k$ ranges over $[\numitems]$.  Note that these estimators
cannot be computed in practice (since the value of $\commstar$ is
unknown), but they are convenient for our analysis, in particular
because $\wtreg = \wtcomm$ for some value $\commsize \in [\numitems]$.

We first show that there exists a universal
constant $\plaincon_0 > 0$ such that
\begin{align}
\label{EqnDahlTadka}
\mprob \Big[ \frobnorm{\wtcomm - \wtstar}^2 > \plaincon_0 (\numitems -
  \commstar + 1) (\log \numitems)^3 \Big] & \leq e^{-(\log
  \numitems)^2}
\end{align}
for each fixed $\commsize \in [\numitems]$.  Since $\wtreg = \wtcomm$
for some $k$, we then have
\begin{align*}
\mprob \Big[ \frobnorm{\wtreg - \wtstar}^2 > \plaincon_0 (\numitems -
  \commstar + 1) (\log \numitems)^3 \Big] & \leq \mprob \Big[
  \max_{\commsize \in [\numitems]} \frobnorm{\wtcomm - \wtstar}^2 >
  \plaincon_0 (\numitems - \commstar + 1) (\log \numitems)^3 \Big] \\
& \stackrel{(i)}{\leq} \numitems e^{-(\log \numitems)^2} \; \leq \;
e^{-\frac{1}{2} (\log \numitems)^2}.
\end{align*}
where step (i) follows from the union bound.  We have thus established
the claimed tail bound~\eqref{EqnRegLseUpper}.

In order to prove the bound~\eqref{EqnRegLseAdaptive} on the
adaptivity index, we first integrate the tail
bound~\eqref{EqnRegLseUpper}.  Since all entries of $\wtstar$ and
$\wtreg$ all lie in $[0,1]$, we have $\frobnorm{\wtstar - \wtreg}^2
\leq \numitems^2$, and so this integration step yields an analogous
bound on the expected error:
\begin{align*}
\Exs [ \frobnorm{ \wtstar - \wtreg}^2 ] \leq \UUP (\numitems -
\maxsizefn(\wtstar) + 1) (\log \numitems)^3.
\end{align*}
Coupled with the lower bound on the risk of the oracle estimator
established in Proposition~\ref{PropORA}, we obtain the claimed
bound~\eqref{EqnRegLseAdaptive} on the adaptivity index of $\wtreg$.

It remains to prove the tail bound~\eqref{EqnDahlTadka}.  We proceed
via a two step argument: first we use the general upper bound given by
Lemma~\ref{LemGeneralUpper} to derive a weaker version of the required
bound; and second, we then refine this weaker bound so as to obtain
the bound~\eqref{EqnDahlTadka}.

\paragraph{Establishing a weaker bound:}

Beginning with the first step, let us apply
Lemma~\ref{LemGeneralUpper} with the choices
\begin{align*}
\bgenup = 0, \quad \classgenup = \chatterjeeclass(\commsize) \cup
\chatterjeeclass(\commstar), \quad \mbox{and} \quad \reggenup(\wt) =
\reg{\maxsizefn(\wt)}.
\end{align*}
With these choices, the $\chattDiff(\wtstar, \ballgenup)$ in the
statement of Lemma~\ref{LemGeneralUpper} takes the form
\begin{align*}
\chattDiff(\wtstar, \ballgenup) \subseteq \{ \alpha(\wt - \wtstar)
\mid \alpha \in [0,1], \wt \in \chatterjeeclass(\commsize) \cup
\chatterjeeclass(\commstar)\}.
\end{align*}
Lemma~\ref{LemChattMetent_adaptive} implies that
\begin{align*}
\metent(\epsilon, \chattDiff(\wtstar, \ballgenup), \frobnorm{.}) &
\leq \plaincon \frac{(\numitems - \min\{\commsize,\commstar\} +
  1)^2}{\epsilon^2} (\log \numitems )^2 + \plaincon (\numitems -
\min\{\commsize,\commstar\} + 1) \log \numitems
\end{align*}
for all $\epsilon \geq \numitems^{-8}$.  Applying
Lemma~\ref{LemGeneralUpper} with $\tailgenup = 1$ then yields
\begin{align}
\label{EqnRegUpWeak}
\mprob \Big( \frobnorm{\wtcomm - \wtstar}^2 > \plaincon (\numitems -
\min\{\commsize,\commstar\} + 1) (\log \numitems )^2 \Big) \leq e^{-
  \plaincon (\log \numitems )^2 }.
\end{align}
Note that this bound is weaker than the desired
bound~\eqref{EqnDahlTadka}, since $\min \{\commsize, \commstar \} \leq
\commstar$.  Thus, our next step is to refine it.

\paragraph{Refining the bound~\eqref{EqnRegUpWeak}:}

Before proceeding with the proof, we must take care of one
subtlety. Recall that the set $\chatterjeeclass(\commstar)$ consists
of all matrices in $\chatterjeeclass$ that have an indifference set
containing at least (but not necessarily exactly) $\commstar$
items. If $\commsize \geq \commstar$, then the
bound~\eqref{EqnRegUpWeak} is equivalent to the
bound~\eqref{EqnDahlTadka}. Otherwise, we evaluate the estimator
$\wtcomm$ for the choices $\commsize = 1, \ldots, \commstar-1$ (in this
particular order).  For any $\commsize \in \{1, \ldots, \commstar-1\}$
under consideration, suppose $\maxsizefn(\wtcomm) = \commsize' <
\commsize$. Then the estimate under consideration is either also an
optimal estimator for the case of $\wt_{\commsize'}$, or it is
suboptimal for the aggregate estimation
problem~\eqref{EqnDefnRegRewrite}. In the former case, the error
incurred by this estimate is already handled in the analysis of
$\wt_{\commsize'}$, and in the latter case, it is
irrelevant. Consequently, it suffices to evaluate the case when
$\maxsizefn(\wtcomm) = \commsize$.

Observe that the matrix $\wtcomm$ is optimal for the optimization
problem~\eqref{EqnDefnRegWeaker} and the matrix $\wtstar$ lies in the
feasible set. Consequently, we have the basic inequality:
\begin{align*}
\frobnorm{\obs - \wtcomm}^2 + \reg{\commsize} \leq \frobnorm{\obs -
  \wtstar}^2 + \reg{\commstar}.
\end{align*}
Using the linearized form of the observation
model~\eqref{EqnLinearized}, some simple algebraic manipulations give
\begin{align}
\label{EqnBasicReg}
\half \frobnorm{ \wtcomm - \wtstar }^2 \leq \tracer{\wtcomm - \wtstar
}{\noise} - \reg{\commsize} + \reg{\commstar},
\end{align}
where $\noise$ is the noise matrix~\eqref{EqnDefnNoise} in the
linearized form of the model. The following lemma helps bound the
first term on the right hand side of
inequality~\eqref{EqnBasicReg}. Consistent with the notation elsewhere
in the paper, for any value of $\ballgenup > 0$, let us define a set
of matrices $\chattDiff(\wtstar, \ballgenup) \subseteq
\chatterjeeclass$ as
\begin{align*}
\chattDiff(\wtstar, \ballgenup) \defn \{ \alpha(\wt - \wtstar) \mid
\wt \in \chatterjeeclass(\commsize),\ \alpha \in
    [0,1],\ \frobnorm{\alpha(\wt - \wtstar)} \leq \ballgenup \}.
\end{align*}
With this notation, we then have the following result:
\begin{lemma}
\label{lem:set_sup_upper}
For any $\wtstar \in \chatterjeeclass$, any fixed $\commsize \in
[\numitems]$, and any $\ballgenup > 0$, we have
\begin{align}
\label{eq:set_sup_upper}
\sup \limits_{\chattDiffmx \in \chattDiff(\wtstar, \ballgenup)}
\tracer{\chattDiffmx }{\noise} \leq \plaincon \ballgenup
\sqrt{(\numitems - \min\{\commsize, \commstar\} + 1)} \log \numitems +
\plaincon(\numitems - \min\{\commsize, \commstar\} + 1) (\log
\numitems)^2
\end{align}
with probability at least $1 - e^{-(\log \numitems)^2}$.
\end{lemma}
\noindent See Section~\ref{SecProofLemSetSupUpper} for the proof of
this lemma.

From our weaker guarantee~\eqref{EqnRegUpWeak}, we know that
$\frobnorm{ \wtcomm - \wtstar } \leq \plaincon' \sqrt{(\numitems -
  \min\{\commsize, \commstar\} + 1) (\log \numitems)^2}$, with high
probability. Consequently, the term $\tracer{\wtcomm - \wtstar
}{\noise}$ is upper bounded by the quantity~\eqref{eq:set_sup_upper}
for some value of $\ballgenup \leq \plaincon' \sqrt{(\numitems -
  \min\{\commsize, \commstar\} + 1) (\log \numitems)^2}$, and hence
\begin{align*}
\tracer{\wtcomm - \wtstar }{\noise} \leq \plaincon'' (\numitems -
\min\{\commsize, \commstar\} + 1) (\log \numitems)^2,
\end{align*}
with probability at least $1 - e^{-(\log \numitems)^2}$. Applying
this bound to the basic inequality~\eqref{EqnBasicReg} and performing
some algebraic manipulations yields the claimed
result~\eqref{EqnDahlTadka}.


\subsubsection{Proof of Lemma~\ref{lem:set_sup_upper}}
\label{SecProofLemSetSupUpper}

Consider the function $\adaptZ: [0,\numitems] \rightarrow \reals_+$
given by \mbox{$\adaptZ(\ballgenup ) \defn \sup \limits_{\chattDiffmx
    \in \chattDiff(\wtstar, \ballgenup) }
  \tracer{\chattDiffmx}{\noise}$.}  In order to control the behavior
of this function, we first bound the metric entropy of the set
$\chattDiff(\wtstar, \ballgenup)$. Note that
Lemma~\ref{LemChattMetent_adaptive} ensures that
\begin{align*}
\metent(\epsilon, \chattDiff(\wtstar, \ballgenup), \frobnorm{\cdot} )
\leq \plaincon \frac{(\numitems - \min\{\commsize, \commstar\} +
  1)^2}{\epsilon^2} \big(\log \frac{\numitems}{\epsilon} \big)^2+
\plaincon(\numitems - \min\{\commsize, \commstar\} + 1) \log
\numitems.
\end{align*}
Based on this metric entropy bound, the truncated version of Dudley's
entropy integral then guarantees that
\begin{align*}
\Exs [ \adaptZ(\ballgenup) ] \leq \plaincon (\numitems -
\min\{\commsize,\commstar\} + 1) (\log \numitems )^2 + \plaincon
\ballgenup \sqrt{(\numitems - \min\{\commsize, \commstar\} + 1) \log
  \numitems}.
\end{align*}
It can be verified that the function $\adaptZ(\ballgenup)$ is
$\ballgenup$-Lipschitz.  Moreover, the random matrix $W$ has entries~\eqref{EqnDefnNoise} that are independent on and above the diagonal, bounded by $1$ in absolute value, and satisfy skew-symmetry. Consequently, Ledoux's concentration theorem~\cite[Theorem
  5.9]{Ledoux01} guarantees that
\begin{align*}
\mprob \big[ \adaptZ(\ballgenup) \geq \Exs[\adaptZ(\ballgenup) ] +
  \ballgenup v \big] \leq e^{-v^2} \qquad \mbox{for all $v \geq 0$.}
\end{align*}
Combining the pieces, we find that
\begin{align*}
\mprob \Big[ \adaptZ(\ballgenup) \geq \plaincon (\numitems -
  \min\{\commsize,\commstar\} + 1) (\log \numitems )^2 + \plaincon
  \ballgenup \sqrt{(\numitems - \min\{\commsize, \commstar\} + 1) \log
    \numitems} + \ballgenup v \Big] \leq e^{-v^2},
\end{align*}
valid for all \mbox{$v \geq 0$.}  Setting $v = \sqrt{(\numitems -
  \min\{\commsize, \commstar\} + 1)} \log \numitems$ yields the
claimed result.


\subsection{Proof of Theorem~\ref{ThmCRLUpper}}

We now prove the upper bound for the CRL estimator, as stated in
Theorem~\ref{ThmCRLUpper}.  In order to simplify the presentation, we
assume without loss of generality that the true permutation of the
$\numitems$ items is the identity permutation $\permid$.  Let
$\permcrl = (\perm_1,\ldots,\perm_\numitems)$ denote the permutation
obtained at the end of the second step of the CRL estimator. The
following lemma proves two useful properties of the outcomes of the
first two steps.

\begin{lemma}
\label{LemStepsOneTwo}
With probability at least $1 - \numitems^{-20}$, the permutation
$\permcrl$ obtained at the end of the first two steps of the estimator
satisfies the following two properties:
\begin{enumerate}[label = (\alph*)]
\item $\max_{i \in [\numitems]} \sum_{\ell = 1}^{\numitems} |
  \wtstar_{i \ell} - \wtstar_{\permcrl(i) \ell} | \leq
   \sqrt{\numitems} (\log \numitems)^2$, and
\item the group of similar items obtained in the first step is of size
  at least $\commstar = \maxsizefn(\wtstar)$.
\end{enumerate}
\end{lemma}
\noindent See Section~\ref{SecProofLemStepsOneTwo} for the proof of
this claim.

Given Lemma~\ref{LemStepsOneTwo}, let us complete the proof of the
theorem.  Let $\permhatclass$ denote the set of all permutations on
$\numitems$ items which satisfy the two conditions (a) and (b) stated
in Lemma~\ref{LemStepsOneTwo}. Given that every entry of $\wtstar$
lies in the interval $[0,1]$, any permutation $\permhat \in
\permhatclass$ satisfies
\begin{align}
\label{EqnRowPermAbs}
\frobnorm{\wtstar - \permhat(\wtstar)}^2 & = \sum_{i \in [\numitems]}
\sum_{\ell \in [\numitems]} (\wtstar_{i \ell} - \wtstar_{\permhat(i)
  \ell} )^2 \; \leq \; \sum_{i \in [\numitems]} \sum_{\ell \in
  [\numitems]} \mid \wtstar_{i \ell} - \wtstar_{\permhat(i) \ell}
\mid.
\end{align}
Now consider any item $i \in [\numitems]$.  Incorrectly estimating
item $i$ as lying in position $\permhat(i)$ contributes a non-zero
error only if either item $i$ or item $\permhat(i)$ lies in the
$(\numitems - \commstar)$-sized set of items outside the largest
indifference set. Consequently, there are at most $2(\numitems -
\commstar)$ values of $i$ in the sum~\eqref{EqnRowPermAbs} that make a
non-zero contribution. Moreover, from property (a) of
Lemma~\ref{LemStepsOneTwo}, each such item contributes at most
$\sqrt{\numitems} (\log \numitems)^2$ to the error. As a consequence,
we have the upper bound
\begin{align}
\label{EqnRowPermFrob}
\frobnorm{\wtstar - \permhat(\wtstar)}^2 & \leq 2 (\numitems -
\commstar) \sqrt{\numitems} (\log \numitems)^2.
\end{align}

Let us now analyze the third step of the CRL estimator. The problem of
bivariate isotonic regression refers to estimation of the matrix
$\wtstar \in \chatterjeeclass$ when the true underlying permutation of
the items \emph{is known} a priori. In our case, the permutation is
known only approximately, so that we need also to track the associated
approximation error.  In order to derive a tail bound on the error of
bivariate isotonic regression, we call upon the general upper bound
proved earlier in Lemma~\ref{LemGeneralUpper} with the choices
$\bgenup = 1$, $\classgenup = \chattperm{\permid}$, and $\reggenup =
0$.  Now let
\begin{align*}
\chattDiff(\wtstar, \ballgenup) \defn \{ \alpha (\wt - \wtstar) \mid
\wt \in \chattperm{\permid} \}.
\end{align*}

The following lemma uses a result from the
paper~\cite{chatterjee2015biisotonic} to derive an upper bound on the
metric entropy of $\chattDiff(\wtstar, \ballgenup)$. For any matrix $\wtstar \in \chatterjeeclass$, let $\numcomm(\wtstar)$ denote the number of indifference sets in $\wtstar$.
\begin{lemma}
\label{LemChattMetent}
For every $\epsilon > \numitems^{-8}$ and $\ballgenup \in (0,
\numitems]$, we have the metric entropy bound
\begin{align*}
\metent(\epsilon, \chattDiff(\wtstar, \ballgenup), \frobnorm{.}) &
\leq \plaincon \frac{\ballgenup^2 (\numcomm(\wtstar))^2 (\log
  \numitems)^6 } {\epsilon^2}.
\end{align*}
where $\plaincon > 0$ is a universal constant.
\end{lemma}

With this bound on the metric entropy, an application of
Lemma~\ref{LemGeneralUpper} with $\tailgenup = \frac{(\numitems - \commstar + 1)^2}{(\numcomm(\wtstar))^2}$ gives that for every
$\wtstar \in \chattperm{\permid}$, the least squares estimator
$\wthat_{\permid} \in \argmin \limits_{\wt \in \chattperm{\permid}}
\frobnorm{\wt - \obs}^2$ incurs an error upper bounded as
\begin{align*}
\frobnorm{\wthat_{\permid} - \wtstar}^2 & \leq \plaincon
(\numitems - \commstar + 1)^2 (\log \numitems)^8,
\end{align*}
with probability at least $1 - e^{- (\numitems - \commstar + 1)^2 (\log \numitems)^8}$. Note that this application of Lemma~\ref{LemGeneralUpper} is valid since $\numcomm(\wtstar) \leq \numitems - \commstar + 1$ and hence $\tailgenup \geq 1$. Furthermore, it
follows from a corollary of Theorem 1 in the
paper~\cite{shah2015stochastically} that
\begin{align*}
\frobnorm{\wthat_{\permid} - \wtstar}^2 \leq \plaincon \numitems (\log
\numitems)^2,
\end{align*}
with probability at least $1 - e^{- \plaincon \numitems}$. Combining
these upper bounds yields
\begin{align}
 \frobnorm{\wthat_{\permid} - \wtstar}^2 & \leq \plaincon
 \min\{(\numitems - \commstar + 1)^2, \numitems\} (\log \numitems)^8 \;
 \stackrel{(i)}{\leq} \plaincon (\numitems - \commstar + 1)
 \sqrt{\numitems} (\log \numitems)^8,
\label{EqnNoiselessBiso}
\end{align}
with probability at least $1 - e^{- \plaincon (\numitems - \commstar + 1)^2 (\log \numitems)^8}$, where
$\plaincon$ is a positive universal constant.  Inequality (i) makes
use of the bound $\min\{u^2,v^2\} \leq uv$ for any two non-negative numbers $u$ and $v$.

Let us put together the analysis of the approximation
error~\eqref{EqnRowPermFrob} in the permutation obtained in the first
two steps and the error~\eqref{EqnNoiselessBiso} in estimating the
matrix in the third step. To this end, consider any permutation
$\permhat \in \permhatclass$. For clarity, we augment the notation of
$\wtCRL$ (defined in~\eqref{EqnDefnCRL}) and use $\wtCRL(\obs,
\permhat)$ to represent the estimator $\wtCRL$ under the permutation
$\permhat$ for the observation matrix $\obs$, that is,
\begin{align*}
\wtCRL(\obs, \permhat) \defn \argmin \limits_{\wt \in
  \chattperm{\permhat}} \frobnorm{\wt - \obs}^2.
\end{align*}
Consider any matrix $\wtstar \in \chattperm{\permid}$ under the
identity permutation.  We can then write
\begin{align}
&\frobnorm{ \wtCRL(\wtstar + \noise, \permhat) - \wtstar}^2 & \nonumber \\
& = \frobnorm{ \wtCRL(\wtstar + \noise, \permhat) -
  \wtCRL(\permhat(\wtstar) + \noise, \permhat) +
  \wtCRL(\permhat(\wtstar) + \noise, \permhat) - \wtstar}^2 \nonumber \\
& \leq 2\frobnorm{ \wtCRL(\wtstar + \noise, \permhat) -
   \wtCRL(\permhat(\wtstar) + \noise, \permhat)}^2 + 2
 \frobnorm{\wtCRL(\permhat(\wtstar) + \noise, \permhat) - \wtstar}^2.
\label{EqnPermcrlTriangle1}
\end{align}
We separately bound the two terms on the right hand side of
equation~\eqref{EqnPermcrlTriangle1}. First observe that the least
squares step of the estimator $\wtCRL$ (for a given permutation
$\permhat$ in its second argument) is a projection onto the convex set
$\chattperm{\permhat}$, and hence we have the deterministic bound
\begin{subequations}
\begin{align}
\label{EqnPermcrlTriangle2}
\frobnorm{ \wtCRL(\wtstar + \noise, \permhat) -
  \wtCRL(\permhat(\wtstar) + \noise, \permhat)}^2 & \leq
\frobnorm{\wtstar - \permhat(\wtstar)}^2 .
\end{align}
In addition, we have
\begin{align}
\label{EqnPermcrlTriangle3}
\frobnorm{\wtCRL(\permhat(\wtstar) + \noise, \permhat) - \wtstar}^2
\leq 2\frobnorm{\wtCRL(\permhat(\wtstar) + \noise, \permhat) -
  \permhat(\wtstar)}^2 + 2 \frobnorm{\permhat(\wtstar) - \wtstar}^2.
\end{align}
\end{subequations}
From our earlier bound~\eqref{EqnNoiselessBiso}, we have that for any
\emph{fixed} permutation $\permhat \in \permhatclass$, the least
squares estimate satisfies
\begin{align}
\label{EqnPermcrlTriangle4}
\frobnorm{\wtCRL(\permhat(\wtstar) + \noise, \permhat) - \permhat(\wtstar)}^2 &
\leq \UUP (\numitems - \commstar + 1) \sqrt{\numitems} (\log
\numitems)^8,
\end{align}
with probability at least $1 - e^{- \plaincon (\numitems - \commstar + 1)^2 (\log \numitems)^8}$.

In conjunction, the
bounds~\eqref{EqnRowPermFrob},~\eqref{EqnPermcrlTriangle1},~\eqref{EqnPermcrlTriangle2},~\eqref{EqnPermcrlTriangle3}
and~\eqref{EqnPermcrlTriangle4} imply that for any \emph{fixed}
$\permhat \in \permhatclass$,
\begin{align}
\label{EqnMarriage}
\mprob \Big( \frobnorm{ & \wtCRL(\wtstar + \noise, \permhat) -
  \wtstar}^2 \leq \UUP (\numitems - \commstar + 1) \sqrt{\numitems}
(\log \numitems)^8 \Big) \geq 1 - e^{-\UHP (\numitems - \commstar + 1)^2
 (\log \numitems)^8}.
\end{align}
Although we are guaranteed that $\permcrl \in \permhatclass$, we
cannot apply the bound~\eqref{EqnMarriage} directly to it, since
$\permcrl$ is a data-dependent quantity.  In order to circumvent this
issue, we need to obtain a uniform version of the
bound~\eqref{EqnMarriage}, and we do so by applying the union bound
over the data-dependent component of $\permcrl$.

In more detail, let us consider Steps $1$ and $2$ of the CRL algorithm
as first obtaining a total ordering of the $\numitems$ items via a
count of the number of pairwise victories, then converting it to a
partial order by putting all items in the subset identified by Step
$2$ in an equivalence class, and then obtaining a total ordering by
permuting the items in the equivalence class in a
\emph{data-independent} manner. Lemma~\ref{LemStepsOneTwo} ensures
that the size of this equivalence class is at least
$\commstar$. Consequently, the number of possible (data-dependent)
partial orders obtained is at most
$\frac{\factorial{\numitems}}{\factorial{\commstar}} \leq e^{
  (\numitems - \commstar) \log \numitems}$. Taking a union bound over
each of these $e^{ (\numitems - \commstar) \log \numitems}$ cases, we
get that
\begin{align*}
\mprob \Big[ \frobnorm{ & \wtCRL(\wtstar + \noise, \permcrl) -
    \wtstar}^2 \leq \UUP (\numitems - \commstar + 1) \sqrt{\numitems}
  (\log \numitems)^8 \mid \permcrl \in \permhatclass \Big ] & \geq 1 -
e^{- (\log \numitems)^7}.
\end{align*}
Recalling that Lemma~\ref{LemStepsOneTwo} ensures that $\mprob \big[
  \permcrl \in \permhatclass \big] \geq 1 - \numitems^{-20}$, we have
established the claim. \\

\noindent It remains to prove the two auxiliary lemmas stated above.


\subsubsection{Proof of Lemma~\ref{LemStepsOneTwo}}
\label{SecProofLemStepsOneTwo} 

We first prove that for any fixed item $i \in [\numitems]$, the
inequality of part (a) holds with probability at least $1 -
\numitems^{-22}$.  The claimed result then follows via a union bound
over all items.

Consider any item $j > i$ such that
\begin{align}
\label{EqnSeparationCondition}
\sum_{\ell=1}^{\numitems} \wtstar_{i \ell} - \sum_{\ell=1}^{\numitems}
\wtstar_{j \ell} > \sqrt{\numitems} (\log \numitems)^2.
\end{align}
An application of the Bernstein inequality then gives (see the proof
of Theorem 1 in the paper~\cite{shah2015simple} for details) that
\begin{align*}
\mprob \big( \sum_{\ell=1}^{\numitems} \obs_{j \ell} \geq
\sum_{\ell=1}^{\numitems} \obs_{i \ell} \big) \leq \frac{1}{\numitems^{23}}.
\end{align*}
Likewise, for any item $j < i$ such that $\sum_{\ell=1}^{\numitems}
\wtstar_{j \ell} - \sum_{\ell=1}^{\numitems} \wtstar_{i \ell} >
 \sqrt{\numitems} (\log \numitems)^2$, we have $\mprob \big(
\sum_{\ell=1}^{\numitems} \obs_{i \ell} \geq \sum_{\ell=1}^{\numitems}
\obs_{j \ell} \big) \leq
\frac{1}{\numitems^{23}}$.

Now consider any $j \geq i$. In order for item $i$ to be located in
position $j$ in the total order given by the row sums, there must be
at least $(j - i)$ items in the set $\{i+1, \ldots, \numitems\}$ whose
row sums are at least as big as the sum of the $i^{th}$ row of
$\obs$. In particular, there must be at least one item in the set
$\{j,\ldots,\numitems\}$ such that its row sum is as big as the sum of
the $i^{th}$ row of $\obs$. It follows from our results above that
under the condition~\eqref{EqnSeparationCondition}, this event occurs
with probability no more than $\frac{1}{\numitems^{21}}$. Likewise
when $j \leq i$, thereby proving the claim.

We now move to the condition of part (b). Observe that for any two
items $i$ and $j$ in the same indifference set, we have that
$\wtstar_{i \ell} = \wtstar_{j \ell}$ for every $\ell \in [\numitems]$. An application of the
Bernstein inequality now gives that
\begin{align*}
\mprob \big( \sum_{\ell=1}^{\numitems} \obs_{j \ell} -
\sum_{\ell=1}^{\numitems} \obs_{i \ell} \geq \sqrt{\numitems} \log
\numitems \big) \leq \frac{1}{\numitems^{23}}.
\end{align*}
A union bound over all pairs of items in the largest indifference set
gives that all $\commstar$ items in the largest indifference set have
their row sums differing from each other by at most $\sqrt{\numitems} \log \numitems$. Consequently, the group must be of
at least this size.


\subsubsection{Proof of Lemma~\ref{LemChattMetent}}

For the proof, it is be convenient to define a class $\bisoclassminus$
that is similar to the class $\chattperm{\permid}$, but contains
matrices with entries in $[-1,1]$:
\begin{align*}
\bisoclassminus{\ballgenup} \defn \big \{ \wt \in [-1,1]^{\numitems
  \times \numitems} \mid \frobnorm{\wt} \leq
\ballgenup,\ \mbox{$\wt_{k \ell} \geq \wt_{ij}$ whenever $k \leq i$
  and $\ell \geq j$} \big \}.
\end{align*}
We now call upon Theorem 3.3 of the
paper~\cite{chatterjee2015biisotonic}.  It provides the following
upper bound on the metric entropy of bivariate isotonic matrices
within a Frobenius ball:
\begin{align*}
\metent ( \epsilon, \bisoclassminus{\ballgenup}, \frobnorm{\cdot} )
\leq \plaincon_0 \frac{t^2 (\log \numitems)^4 } {\epsilon^2} \big(
\log \frac{t \log \numitems}{\epsilon} \big)^2.
\end{align*}
Substituting $\epsilon \geq \numitems^{-8}$ and $\ballgenup \leq
\numitems$ yields
\begin{align}
\label{EqnDefnMetentBisoMinus}
\metent ( \epsilon, \bisoclassminus{\ballgenup}, \frobnorm{\cdot} )
\leq \plaincon \frac{t^2 (\log \numitems)^6 } {\epsilon^2}.
\end{align}

We now use this result to derive an upper bound on the metric entropy
of the set $ \chattDiff(\wtstar, \ballgenup)$. Consider the following
partition of the entries of any $(\numitems \times \numitems)$ matrix
into $(\numcomm(\wtstar))^2$ submatrices. Submatrix $(i,j) \in
[\numcomm(\wtstar)] \times [\numcomm(\wtstar)]$ in this partition is
the $(\commsize_i \times \commsize_j)$ submatrix corresponding to the
pairwise comparison probabilities between every item in the $i^{th}$
indifference set with every item in the $j^{th}$ indifference set in
$\wtstar$. Such a partition ensures that each partitioned submatrix of
$\wtstar$ is a constant matrix. Consequently, for any $\wt \in
\chattDiff(\wtstar, \ballgenup)$, each partitioned submatrix belongs
to the set of matrices $\bisoclassminus{\ballgenup}$ (where we
slightly abuse notation to ignore the size of the matrices as long as
no dimension is greater than $(\numitems \times \numitems)$). The
metric entropy of the set of matrices in $\chattDiff(\wtstar,
\ballgenup)$ can now be upper bounded by the sum of the metric
entropies of each set of submatrices. Consequently, we have
\begin{align*}
\metent(\epsilon, \chattDiff(\wtstar, \ballgenup), \frobnorm{\cdot}) &
\leq (\numcomm(\wtstar))^2 \metent(\epsilon,
\bisoclassminus{\ballgenup} , \frobnorm{\cdot}) \\
& \leq \plaincon \frac{t^2 (\numcomm(\wtstar))^2 (\log \numitems)^6 }
               {\epsilon^2},
\end{align*}
where the final inequality follows from our earlier
bound~\eqref{EqnDefnMetentBisoMinus}.


\subsection{Proof of Theorem~\ref{ThmPolyLower}}

We now turn to the proof of the lower bound for polynomial-time
computable estimators, as stated in Theorem~\ref{ThmPolyLower}.  We
proceed via a reduction argument.  Consider any estimator that has
Frobenius norm error upper bounded as
\begin{align}
\label{EqnLowPoly1}
\sup_{\wtstar \in \chatterjeeclass} & \mathbb{E}
    [\frobnorm{\wthat - \wtstar}^2 ] \leq \UUP \sqrt{n} (\numitems -
    \maxsizefn(\wtstar) + 1) (\log \numitems)^{-1}.
\end{align}
We show that any such estimator defines a method that, with
probability at least $1 - \frac{1}{\sqrt{\log \numitems}}$, is able to
identify the presence or absence a planted clique with
$\frac{\sqrt{\numitems}}{\log \log \numitems}$ vertices.  This result,
coupled with the upper bound on the risk of the oracle estimator
established in Proposition~\ref{PropORA} proves the claim of
Theorem~\ref{ThmPolyLower}.

Our reduction from the bound~\eqref{EqnLowPoly1} proceeds by
identifying a subclass of $\chatterjeeclass$, and showing that any
estimator satisfying the bound~\eqref{EqnLowPoly1} on on this subclass
can be used to identify a planted clique in an Erd\H{o}s-R\'enyi
random graph.  Naturally, in order to leverage the planted clique
conjecture, we need the planted clique to be of size $o(\sqrt{n})$.

Our construction involves a partition with $\numcomm = 3$ components,
maximum indifference set size $\commmax = \commsize_1 = \numitems -
2\commsize$, with the remaining two indifference sets of size
$\commsize_2 = \commsize_3 = \commsize$.  We choose the parameter
$\commsize \defn \frac{\sqrt{\numitems}}{\log \log \numitems}$ so that
any constant multiple of it will be within the hardness regime of
planted clique (for sufficiently large values of $\numitems$). Now let
$\wtstar_0$ be a matrix with all ones in the $(\commsize \times \commsize)$
sub-matrix in its top-right, zeros on the corresponding sub-matrix in
the bottom-left and all other entries set equal to $\half$.  By
construction, the matrix $\wtstar_0$ belongs to the class
$\chatterjeeclass(\commmax)$ with $\commmax = \numitems - 2
\commsize$.

For any permutation $\perm$ on $\frac{\numitems}{2}$ items and any
$(\numitems \times \numitems)$ matrix $\wt$, define another
$(\numitems \times \numitems)$ matrix $\HACKPERM(\wt)$ by applying the
permutation $\perm$ to:

\begin{itemize}[leftmargin = *]
\item the first $\frac{\numitems}{2}$ rows of $\wtstar$, and the last
  $\frac{\numitems}{2}$ rows of $\wtstar$
\item the first $\frac{\numitems}{2}$ columns of $\wtstar$, and  to the
  last $\frac{\numitems}{2}$ columns of $\wtstar$.
\end{itemize}
We then define the set $\chatterjeeclassPC \defn \Big \{
\HACKPERM(\wtstar_0) \mid \mbox{for all permutations $\perm$ on
  $[\numitems/2]$} \Big \}$.  By construction, it is a subset of
$\chatterjeeclass(\numitems - \commmax)$.

 For any estimator $\wthat$ that satisfies the
bound~\eqref{EqnLowPoly1}, we have
\begin{align*}
\sup_{\wtstar \in \chatterjeeclassPC \cup \{\frac{1}{2} \ones \ones^T
  \}} \Exs [ \frobnorm{\wthat - \wtstar}^2 ] \leq \plaincon \commsize
\frac{\sqrt{\numitems}}{\log \numitems}.
\end{align*}
On the other hand, Markov's inequality implies that
\begin{align*}
\Exs [ \frobnorm{\wthat - \wtstar}^2 ] & > \plaincon \frac{\commsize
  \sqrt{\numitems}}{\sqrt{\log \numitems}} \mprob \Big [
  \frobnorm{\wthat - \wtstar}^2 \geq \plaincon \frac{\commsize
    \sqrt{\numitems}}{\sqrt{\log \numitems}} \Big].
\end{align*}
Combining the two bounds, we find that
\begin{align}
\label{eq:upper_planted}
\mprob \Big[ \frobnorm{\wthat - \wtstar}^2 < \frac{\plaincon \,
    \commsize \sqrt{\numitems}}{\sqrt{\log \numitems}} \Big] \geq 1 -
\frac{1}{\sqrt{\log \numitems}}.
\end{align}

Consider the set of $(\frac{\numitems}{2} \times \frac{\numitems}{2})$
matrices comprising the top-right $(\frac{\numitems}{2} \times
\frac{\numitems}{2})$ sub-matrix of every matrix in
$\chatterjeeclassPC$. We claim that this set is identical to the set of all possible
matrices in the planted clique problem with $\frac{\numitems}{2}$
vertices and a planted clique of size $\commsize$. Indeed, the set contains the all-half matrix corresponding to the absence of a planted clique, and all symmetric matrices that have all entries equal to half except for a $(k \times k)$ all-ones submatrix corresponding to the planted clique.

Now consider the problem of testing the hypotheses of whether
$\wtstar$ is equal to the all-half matrix (``no planted clique'') or
if it lies in $\chatterjeeclassPC$ (``planted clique''). Let us
consider a decision rule that declares the absence of a planted clique
if $\frobnorm{\wthat - \half \ones \ones^T}^2 \leq \frac{1}{16}
\commsize^2$, and the presence of a planted clique otherwise. 

\paragraph{Null case:}  On one hand, if
there is no planted clique ($\wtstar = \half \ones \ones^T$), then the
bound~\eqref{eq:upper_planted} guarantees that
\begin{align}
\label{eq:upper_planted2}
\frobnorm{\wthat - \half \ones \ones^T}^2 < \commsize
\frac{\sqrt{\numitems}}{\sqrt{\log \numitems}}
\end{align}
with probability at least $1 - \frac{1}{\sqrt{\log
    \numitems}}$. Recalling that $\commsize =
\frac{\sqrt{\numitems}}{\log \log \numitems}$, we find that our
decision rule can detect the absence of the planted clique with
probability at least $1 - \frac{1}{\sqrt{\log \numitems}}$. 

\paragraph{Case of planted clique:} On the
other hand, if there is a planted clique ($\wtstar \in
\chatterjeeclassPC$), then we have
\begin{align*}
\frobnorm{\wthat - \half \ones \ones^T}^2 \geq \half \frobnorm{\wtstar -
  \half \ones \ones^T}^2 - \frobnorm{\wthat - \wtstar}^2 = \frac{1}{4}
\commsize^2 - \frobnorm{\wthat - \wtstar}^2.
\end{align*}
Thus, in this case, the bound~\eqref{eq:upper_planted} guarantees
that
\begin{align*}
\frobnorm{\wthat - \half \ones \ones^T}^2 & \geq \frac{1}{4} \commsize^2 -
\frac{\commsize \sqrt{\numitems}}{\sqrt{\log \numitems}},
\end{align*}
with probability at least $1 - \frac{1}{\sqrt{\log \numitems}}$.
Since $\commsize = \frac{\sqrt{\numitems}}{\log \log \numitems}$, our
decision rule successfully detects the presence of a planted clique
with probability at least $1 - \frac{1}{\sqrt{\log \numitems}}$. 

In summary, given the planted clique conjecture, our decision rule
cannot be computed in polynomial time.  Since it can be computed in
polynomial-time given the estimator $\wthat$, it must also be the case
that $\wthat$ cannot be computed in polynomial time, as claimed.


\subsection{Proof of Theorem~\ref{thm:break_lse}}
\label{SecProofBreakLSE}
We now prove lower bounds on the standard least-squares estimator.  A
central piece in our proof is the following lemma, which characterizes
an interesting structural property of the least-squares estimator.
\begin{lemma}
\label{LemLSTech}
Let $\wtstar = \half \ones \ones^T$ and consider any matrix $\obs \in
\{0,1\}^{\numitems \times \numitems}$ satisfying the
shifted-skew-symmetry condition. Then the least squares estimator
$\wtlse$ from equation~\eqref{EqnDefnLSE} must satisfy the quadratic
equation
\begin{align*}
\frobnorm{\obs - \wtstar}^2 &= \frobnorm{\obs - \wtlse}^2 +
\frobnorm{\wtstar - \wtlse}^2.
\end{align*}
\end{lemma}
\noindent See Section~\ref{SecProofLemTSTech} for the proof of this
claim.\\

Let us now complete the proof of Theorem~\ref{thm:break_lse} using
Lemma~\ref{LemLSTech}.  Our strategy is as follows: we first construct
a ``bad'' matrix $\wttil \in \chatterjeeclass$ that is far from
$\wtstar$ but close to $\obs$.  We then use Lemma~\ref{LemLSTech} to
show that the least squares estimate $\wtlse$ must also be far from
$\wtstar$.

In the matrix $\obs$, let item $\ell$ be an item that has won the
maximum number of pairwise comparisons---that is $\ell \in \argmax
\limits_{i \in [\numitems]} \sum_{j =1}^{\numitems} \obs_{j i}$.  Let
$S$ denote the set of all items that are beaten by item $\ell$---that
is, $S \defn \{ j \in [\numitems] \backslash \{\ell\} \mid \obs_{\ell
  j} = 1\}$.  Note that $\card(S) \geq \frac{\numitems - 1}{2}$.  Now
define a matrix $\wttil \in \chatterjeeclass$ with entries
$\wttil_{\ell,j} = 1 = 1 - \wttil_{j,\ell}$ for every $j \in S$, and
all remaining entries equal to $\half$. Some simple calculations then
give
\begin{subequations}
\begin{align}
\frobnorm{\obs - \wtstar}^2 &= \frobnorm{\obs - \wttil}^2 +
\frobnorm{\wtstar - \wttil}^2, \qquad \mbox{and}\\
\frobnorm{\wttil - \wtstar}^2 &\geq \frac{\numitems - 1}{4}.
\end{align}
\label{eq:wttil_properties_breaklse}
\end{subequations}

Next we exploit the structural property of the least squares solution
guaranteed by Lemma~\ref{LemLSTech}.  Together with the
conditions~\eqref{eq:wttil_properties_breaklse} and the fact that
$\frobnorm{\obs - \wtlse}^2 \leq \frobnorm{\obs - \wttil}^2$, some
simple algebraic manipulations yield the lower bound
\begin{align}
\label{EqnGujurat}
\frobnorm{\wtstar - \wtlse}^2 \geq \frac{\numitems - 1}{4}.
\end{align}
This result holds for any arbitrary observation matrix $\obs$, and
consequently, holds with probability $1$ when the observation matrix
$\obs$ is drawn at random.  For $\commax = \numitems - 1$,
Proposition~\ref{PropORA} yields an upper bound of $\plaincon (\log
\numitems)^{2}$ on the oracle risk.  Combining this upper bound with
the lower bound~\eqref{EqnGujurat} yields the claimed lower bound on
the adaptivity index of the least squares estimator.


\subsubsection{Proof of Lemma~\ref{LemLSTech}}
\label{SecProofLemTSTech}

From our earlier construction of $\wttil$ in Section~\ref{SecProofBreakLSE}, we know that
$\frobnorm{\obs - \wtlse} \leq \frobnorm{\obs - \wttil} <
\frobnorm{\obs - \wtstar}$, which guarantees that $\wtlse \neq
\wtstar$.  Consequently, we may consider the line 
\begin{align*}
\LINE \defn \{\theta \wtstar + (1 - \theta) \wtlse \mid \theta \in
\reals \}
\end{align*}
that passes through the two points $\wtstar$ and $\wtlse$.  Given this
line, consider the auxiliary estimator
\begin{align}
\label{eq:LemLSELine_opt_relax}
\wthat_1 & \defn \argmin \limits_{\wt \in \LINE} \frobnorm{\obs -
  \wt}^2.
\end{align}
Since $\wthat_1$ is the Euclidean projection of $\obs$ onto this line,
it must satisfy the Pythagorean relation
\begin{align}
\frobnorm{\obs - \wtstar}^2 &= \frobnorm{\obs - \wthat_1}^2 +
\frobnorm{\wtstar - \wthat_1}^2.
\label{eq:LemLSELine_pythagorus}
\end{align}
Let $\projunit: \reals^{\numitems \times \numitems} \rightarrow
[0,1]^{\numitems \times \numitems}$ denote the Euclidean projection of
any $(\numitems \times \numitems)$ matrix onto the hypercube
$[0,1]^{\numitems \times \numitems}$. This projection actually has a
simple closed-form expression: it simply clips every entry of the
matrix $\wt$ to lie in the unit interval $[0,1]$. Since projection
onto the convex set $[0,1]^{\numitems \times \numitems}$ is
non-expansive, we must have
\begin{align}
\label{eq:LemLSELine_proj}
\frobnorm{\obs - \wthat_1}^2 \geq \frobnorm{\projunit(\obs) -
  \projunit(\wthat_1)}^2 = \frobnorm{\obs - \projunit(\wthat_1)}^2.
\end{align}
Here the final equation follows since $\obs \in [0,1]^{\numitems
  \times \numitems}$, and hence $\projunit(\obs) = \obs$.

Furthermore, we claim that $\projunit(\wthat_1) \in \chatterjeeclass$.
In order to prove this claim, first recall that the matrix $\wthat_1$
can be written as $\wthat_1 = (1 - \theta) (\wtlse - \half \ones
\ones^T) + \half \ones \ones^T$ for some $\theta \in \reals$, and
$\wtlse \in \chatterjeeclass$.  Consequently, if $\theta \leq 1$, then
the rows/columns of the projected matrix $\projunit(\wthat_1)$ obey
the same monotonicity conditions as those of $\wtlse$; conversely, if
$\theta > 1$, the rows/columns obey an inverted set of monotonicity
conditions, again specified by the rows/columns of $\wthat_1$.
Moreover, since the two matrices $\wtlse$ and $\half \ones \ones^T$
satisfy shifted-skew-symmetry, so does the matrix $\wthat_1$. One can
further verify that any two real numbers $a \geq b$ must also satisfy
the inequalities
\begin{align*}
\min(a,1) \geq \min(b,1), \quad \mbox{and} \quad \max(a,0) \geq \max(b,0).
\end{align*}
If in addition, the pair $(a,b)$ satisfy the constraint, $a + b = 1$,
then we have $\max(\min(a,1),0) + \max(\min(b,1),0) = 1$. Using these
elementary facts, it can be verified that the monotonicity and
shifted-skew-symmetry conditions of any matrix are thus retained by
the projection $\projunit$.

The arguments above imply that $\projunit(\wthat_1) \in
\chatterjeeclass$ and hence the matrix $\projunit(\wthat_1)$ is
feasible for the optimization problem~\eqref{EqnDefnLSE}. By the
optimality of $\wtlse$, we must have 
$\frobnorm{\obs - \projunit(\wthat_1)}^2 \geq \frobnorm{\obs -
  \wtlse}^2$.
Coupled with the inequality~\eqref{eq:LemLSELine_proj}, we find that
\begin{align*}
\label{EqnNanOne}
\frobnorm{\obs - \wthat_1}^2 \geq \frobnorm{\obs - \wtlse}^2.
\end{align*}

On the other hand, since $\wtlse$ is feasible for the optimization
problem~\eqref{eq:LemLSELine_opt_relax} and $\wthat_1$ is the optimal
solution, we must actually have
\begin{align*}
\frobnorm{\obs - \wthat_1}^2 = \frobnorm{\obs - \wtlse}^2,
\end{align*}
so that $\wtlse$ is also optimal for the optimization
problem~\eqref{eq:LemLSELine_opt_relax}.  However, the optimization
problem~\eqref{eq:LemLSELine_opt_relax} amounts to Euclidean
projection on to a line, it must have a unique minimizer, which
implies that $\wtlse = \wthat_1$. Substituting this condition in the
Pythagorean relation~\eqref{eq:LemLSELine_pythagorus} yields the
claimed result.


\section{Conclusions}
\label{SecDiscussion}

We proposed the notion of an adaptivity index to measure the abilities
of any estimator to automatically adapt to the intrinsic complexity of
the problem. This notion helps to obtain a more nuanced evaluation of
any estimator that is more informative than the classical notion of
the worst-case error. We provided sharp characterizations of the
optimal adaptivity that can be achieved in a statistical
(information-theoretic) sense, and that can be achieved by
computationally efficient estimators.  

The logarithmic factors in our results arise from corresponding
logarithmic factors in the metric entropy results of Gao and
Wellner~\cite{gao2007entropy}, and understanding their necessity is an
open question.  In statistical practice, we often desire estimators,
that perform well in a variety of different senses. We believe that
estimating SST matrices at the minimax-optimal rate in Frobenius norm,
as studied in more detail in the paper~\cite{shah2015stochastically},
is also computationally difficult.  We hope to formally establish this
in future work.  Finally, developing a broader understanding of
fundamental limits imposed by computational considerations in
statistical problems is an important avenue for continued
investigation.


\noindent {\emph{Acknowledgements:}} This work was partially supported
by NSF grant CIF-31712-23800, ONR-MURI grant DOD 002888, and AFOSR
grant FA9550-14-1-0016. The work of NBS was supported in part by a
Microsoft Research PhD fellowship.

\confversion{ \bibliographystyle{IEEEtran} } \fullversion{
  \bibliographystyle{alpha_initials} } \bibliography{bibtex}

\newcommand{\etalchar}[1]{$^{#1}$}
\begin{thebibliography}{RWDR88}

\bibitem[AAK{\etalchar{+}}07]{alon2007testing_short}
N. Alon, A. Andoni, T. Kaufman, K. Matulef, R. Rubinfeld, and N. Xie.
\newblock Testing k-wise and almost k-wise independence.
\newblock In {\em ACM STOC}, 2007.

\bibitem[AS15]{abbe2015recovering}
E. Abbe and C. Sandon.
\newblock Recovering communities in the general stochastic block model without
  knowing the parameters.
\newblock In {\em Advances in Neural Information Processing Systems}, pages
  676--684, 2015.

\bibitem[BBM05]{Bar05}
P.~L. Bartlett, O. Bousquet, and S. Mendelson.
\newblock Local {R}ademacher complexities.
\newblock {\em Annals of {S}tatistics}, 33(4):1497--1537, 2005.

\bibitem[BDPR84]{bril1984algorithm}
G. Bril, R. Dykstra, C. Pillers, and T. Robertson.
\newblock Algorithm as 206: isotonic regression in two independent variables.
\newblock {\em Journal of the Royal Statistical Society. Series C (Applied
  Statistics)}, 33(3):352--357, 1984.

\bibitem[Bel16]{bellec2016adaptive}
P.~C. Bellec.
\newblock Adaptive confidence sets in shape restricted regression.
\newblock {\em arXiv preprint arXiv:1601.05766}, 2016.

\bibitem[BM08]{braverman2008noisy}
M. Braverman and E. Mossel.
\newblock Noisy sorting without resampling.
\newblock In {\em Proc. ACM-SIAM symposium on Discrete algorithms}, pages
  268--276, 2008.

\bibitem[BR13]{rigollet}
Q. Berthet and P. Rigollet.
\newblock Complexity theoretic lower bounds for sparse principal component
  detection.
\newblock In {\em {COLT} 2013 - The 26th Annual Conference on Learning Theory,
  June 12-14, 2013, Princeton University, NJ, {USA}}, 2013.

\bibitem[BT52]{bradley1952rank}
R.~A. Bradley and M.~E. Terry.
\newblock Rank analysis of incomplete block designs: I. {T}he method of paired
  comparisons.
\newblock {\em Biometrika}, pages 324--345, 1952.

\bibitem[BW97]{ballinger1997decisions}
T.~P. Ballinger and N.~T. Wilcox.
\newblock Decisions, error and heterogeneity.
\newblock {\em The Economic Journal}, 107(443):1090--1105, 1997.

\bibitem[C{\etalchar{+}}11]{cator2011adaptivity}
E. Cator et~al.
\newblock Adaptivity and optimality of the monotone least-squares estimator.
\newblock {\em Bernoulli}, 17(2):714--735, 2011.

\bibitem[Can06]{candes2006modern}
E.~J. Candes.
\newblock Modern statistical estimation via oracle inequalities.
\newblock {\em Acta numerica}, 15:257--325, 2006.

\bibitem[CGS13]{chatterjee2013risk}
S. Chatterjee, A. Guntuboyina, and B. Sen.
\newblock On risk bounds in isotonic and other shape restricted regression
  problems.
\newblock {\em arXiv preprint arXiv:1311.3765}, 2013.

\bibitem[CGS15]{chatterjee2015biisotonic}
S. Chatterjee, A. Guntuboyina, and B. Sen.
\newblock On matrix estimation under monotonicity constraints.
\newblock {\em \href{http://arxiv.org/abs/1506.03430}{arXiv:1506.03430}}, 2015.

\bibitem[Cha14]{chatterjee2014matrix}
S. Chatterjee.
\newblock Matrix estimation by universal singular value thresholding.
\newblock {\em The Annals of Statistics}, 43(1):177--214, 2014.

\bibitem[CL11]{cai2011framework}
T.~T. Cai and M.~G. Low.
\newblock A framework for estimation of convex functions.
\newblock Technical report, Technical report, 2011.

\bibitem[CL15]{chatterjee2015adaptive}
S. Chatterjee and J. Lafferty.
\newblock Adaptive risk bounds in unimodal regression.
\newblock {\em arXiv preprint arXiv:1512.02956}, 2015.

\bibitem[DJKP95]{donoho1995wavelet}
D.~L. Donoho, I.~M. Johnstone, G. Kerkyacharian, and D. Picard.
\newblock Wavelet shrinkage: asymptopia?
\newblock {\em Journal of the Royal Statistical Society. Series B
  (Methodological)}, pages 301--369, 1995.

\bibitem[DM59]{davidson1959experimental}
D. Davidson and J. Marschak.
\newblock Experimental tests of a stochastic decision theory.
\newblock {\em Measurement: Definitions and theories}, pages 233--69, 1959.

\bibitem[DM15]{deshpande2015improved}
Y. Deshpande and A. Montanari.
\newblock Improved sum-of-squares lower bounds for hidden clique and hidden
  submatrix problems.
\newblock {\em \href{http://arxiv.org/abs/1502.06590}{arXiv:1502.06590}}, 2015.

\bibitem[Dug14]{dughmi2014hardness}
S. Dughmi.
\newblock On the hardness of signaling.
\newblock In {\em IEEE Foundations of Computer Science (FOCS)}, pages 354--363,
  2014.

\bibitem[FK03]{feige2003probable}
U. Feige and R. Krauthgamer.
\newblock The probable value of the lov{\'a}sz--schrijver relaxations for
  maximum independent set.
\newblock {\em SIAM Journal on Computing}, 32(2):345--370, 2003.

\bibitem[Gil52]{gilbert1952comparison}
E.~N. Gilbert.
\newblock A comparison of signalling alphabets.
\newblock {\em Bell System Technical Journal}, 31(3):504--522, 1952.

\bibitem[GW07]{gao2007entropy}
F. Gao and J.~A. Wellner.
\newblock Entropy estimate for high-dimensional monotonic functions.
\newblock {\em Journal of Multivariate Analysis}, 98(9):1751--1764, 2007.

\bibitem[HMG07]{herbrich2007trueskill}
R. Herbrich, T. Minka, and T. Graepel.
\newblock Trueskill: A {B}ayesian skill rating system.
\newblock In {\em Advances in Neural Information Processing Systems}, 2007.

\bibitem[HOX14]{hajek2014minimax}
B. Hajek, S. Oh, and J. Xu.
\newblock Minimax-optimal inference from partial rankings.
\newblock In {\em Advances in Neural Information Processing Systems}, pages
  1475--1483, 2014.

\bibitem[Jer92]{jerrum1992large}
M. Jerrum.
\newblock Large cliques elude the metropolis process.
\newblock {\em Random Structures \& Algorithms}, 3(4):347--359, 1992.

\bibitem[JP00]{juels2000hiding}
A. Juels and M. Peinado.
\newblock Hiding cliques for cryptographic security.
\newblock {\em Designs, Codes and Cryptography}, 20(3):269--280, 2000.

\bibitem[Kol06]{Kolt06}
V. Koltchinskii.
\newblock Local {R}ademacher complexities and oracle inequalities in risk
  minimization.
\newblock {\em Annals of {S}tatistics}, 34(6):2593--2656, 2006.

\bibitem[Kol11]{koltchinskii2011oracle}
V. Koltchinskii.
\newblock {\em Oracle Inequalities in Empirical Risk Minimization and Sparse
  Recovery Problems}, volume~38.
\newblock Springer Science \& Business Media, 2011.

\bibitem[KRS15]{kyng2015fast}
R. Kyng, A. Rao, and S. Sachdeva.
\newblock Fast, provable algorithms for isotonic regression in all l\_p-norms.
\newblock In {\em Advances in Neural Information Processing Systems}, pages
  2701--2709, 2015.

\bibitem[Ku{\v{c}}95]{kuvcera1995expected}
L. Ku{\v{c}}era.
\newblock Expected complexity of graph partitioning problems.
\newblock {\em Discrete Applied Mathematics}, 57(2):193--212, 1995.

\bibitem[Led01]{Ledoux01}
M. Ledoux.
\newblock {\em The {C}oncentration of {M}easure {P}henomenon}.
\newblock Mathematical Surveys and Monographs. American Mathematical Society,
  Providence, RI, 2001.

\bibitem[Luc59]{luce1959individual}
R.~D. Luce.
\newblock {\em Individual choice behavior: A theoretical analysis}.
\newblock New York: Wiley, 1959.

\bibitem[ML65]{mclaughlin1965stochastic}
D.~H. McLaughlin and R.~D. Luce.
\newblock Stochastic transitivity and cancellation of preferences between
  bitter-sweet solutions.
\newblock {\em Psychonomic Science}, 2(1-12):89--90, 1965.

\bibitem[MPW15]{meka2015sum}
R. Meka, A. Potechin, and A. Wigderson.
\newblock Sum-of-squares lower bounds for planted clique.
\newblock {\em \href{http://arxiv.org/abs/1503.06447}{arXiv:1503.06447}}, 2015.

\bibitem[MW15]{ma2015computational}
Z. Ma and Y. Wu.
\newblock Computational barriers in minimax submatrix detection.
\newblock {\em The Annals of Statistics}, 43(3):1089--1116, 2015.

\bibitem[NOS12]{negahban2012iterative}
S. Negahban, S. Oh, and D. Shah.
\newblock Iterative ranking from pair-wise comparisons.
\newblock In {\em Advances in Neural Information Processing Systems}, pages
  2474--2482, 2012.

\bibitem[RGLA15]{rajkumar2015ranking}
A. Rajkumar, S. Ghoshal, L.-H. Lim, and S. Agarwal.
\newblock Ranking from stochastic pairwise preferences: Recovering {Condorcet}
  winners and tournament solution sets at the top.
\newblock In {\em International Conference on Machine Learning}, 2015.

\bibitem[RKJ08]{radlinski2008does}
F. Radlinski, M. Kurup, and T. Joachims.
\newblock How does clickthrough data reflect retrieval quality?
\newblock In {\em ACM conference on Information and knowledge management},
  pages 43--52, 2008.

\bibitem[RWDR88]{robertson1988order}
T. Robertson, F. Wright, R.~L. Dykstra, and T. Robertson.
\newblock {\em Order restricted statistical inference}, volume 229.
\newblock Wiley New York, 1988.

\bibitem[SBB{\etalchar{+}}15]{shah2015estimation}
N.~B. Shah, S. Balakrishnan, J. Bradley, A. Parekh, K. Ramchandran, and M.
  Wainwright.
\newblock Estimation from pairwise comparisons: Sharp minimax bounds with
  topology dependence.
\newblock In {\em Conference on Artificial Intelligence and Statistics}, pages
  856--865, 2015.

\bibitem[SBGW15]{shah2015stochastically}
N.~B. Shah, S. Balakrishnan, A. Guntuboyina, and M.~J. Wainwright.
\newblock Stochastically transitive models for pairwise comparisons:
  Statistical and computational issues.
\newblock {\em arXiv preprint 1510.05610}, 2015.

\bibitem[SW15]{shah2015simple}
N.~B. Shah and M.~J. Wainwright.
\newblock Simple, robust and optimal ranking from pairwise comparisons.
\newblock {\em arXiv preprint 1512.08949}, 2015.

\bibitem[THT11]{tibshirani2011nearly}
R.~J. Tibshirani, H. Hoefling, and R. Tibshirani.
\newblock Nearly-isotonic regression.
\newblock {\em Technometrics}, 53(1):54--61, 2011.

\bibitem[Thu27]{thurstone1927law}
L.~L. Thurstone.
\newblock A law of comparative judgment.
\newblock {\em Psychological Review}, 34(4):273, 1927.

\bibitem[Var57]{varshamov1957estimate}
R. Varshamov.
\newblock Estimate of the number of signals in error correcting codes.
\newblock In {\em Dokl. Akad. Nauk SSSR}, 1957.

\bibitem[vdG00]{van2000empirical}
S. van~de Geer.
\newblock {\em Empirical Processes in {M}-estimation}, volume~6.
\newblock Cambridge university press, 2000.

\bibitem[Wai14]{Wai14_ICM}
M.~J. Wainwright.
\newblock Constrained forms of statistical minimax: {C}omputation,
  communication and privacy.
\newblock In {\em Proceedings of the {I}nternational {C}ongress of
  {M}athematicians}, Seoul, Korea, 2014.

\end{thebibliography}


\end{document}